\PassOptionsToPackage{table}{xcolor}
\documentclass[10pt,twocolumn,letterpaper]{article}

\usepackage[final]{cvpr} 
\usepackage{gensymb}
\usepackage{graphicx}
\usepackage{mathtools} 
\usepackage{amsmath}
\usepackage{amssymb}
\usepackage{booktabs}
\usepackage{dsfont}
\usepackage{multirow}
\usepackage{multicol}
\usepackage{makecell}
\usepackage[pagebackref,breaklinks,colorlinks]{hyperref}
\usepackage[usenames,dvipsnames]{xcolor}
\usepackage{tikz}
\usepackage{pgfplots}
\pgfplotsset{compat=1.18}

\usepackage{pgfkeys}
\usepackage{times}
\usepackage{epsfig}
\usepackage{wrapfig}
\usepackage{comment}
\usepackage{adjustbox}
\usepackage{caption}
\usepackage{microtype}
\usepackage{array}
\usepackage[disable]{todonotes}
\usepackage{pifont}
\usepackage{xcolor}
\usepackage{arydshln}  
\usepackage[font=small,labelsep=period]{caption}
\usepackage{enumitem}
\usepackage{xspace}
\usepackage[normalem]{ulem}

\usepackage{tikz}
\usepackage[accsupp]{axessibility}

\usetikzlibrary{matrix}

\newenvironment{customlegend}[1][]{%
    \begingroup
    \csname pgfplots@init@cleared@structures\endcsname
    \pgfplotsset{#1}%
}{%
    \csname pgfplots@createlegend\endcsname
    \endgroup
}%
\def\addlegendimage{\csname pgfplots@addlegendimage\endcsname}

\usepackage[capitalize]{cleveref}
\crefname{section}{Sec.}{Secs.}
\Crefname{section}{Section}{Sections}
\Crefname{table}{Table}{Tables}
\crefname{table}{Tab.}{Tabs.}

\usepackage{dsfont}

\newif\ifshowedits

\newcommand{\addeditor}[3]{%
  \definecolor{#1color}{rgb}{#3}
  \expandafter\newcommand\csname #1\endcsname[1]{%
  \ifshowedits
    {\color{#1color} ##1}%
  \else
    {##1}%
  \fi
  }%
  \expandafter\newcommand\csname #1rmk\endcsname[1]{%
  \ifshowedits
    {\color{#1color} {\bf [#2: ##1]}}
  \fi
  }%
  \expandafter\newcommand\csname #1rpl\endcsname[2]{%
  \ifshowedits
    {{\color{#1color} ##1} \sout{##2}}
  \else
    {##1}
  \fi
  }%
}

\newcommand{\mycomment}[1]{}

\newcommand{\calL}{{\cal L}}

\newcommand{\calQ}{{\cal Q}}

\newcommand{\calT}{{\cal T}}

\newcommand{\bq}{{\bf q}}

\newcommand{\bt}{{\bf t}}

\newcommand{\bM}{{\bf M}}

\newcommand{\bR}{{\bf R}}

\DeclareMathOperator*{\argmax}{arg\,max}

\DeclareFontFamily{U}{mathx}{\hyphenchar\font45}
\DeclareFontShape{U}{mathx}{m}{n}{
      <5> <6> <7> <8> <9> <10>
      <10.95> <12> <14.4> <17.28> <20.74> <24.88>
      mathx10
      }{}
\DeclareSymbolFont{mathx}{U}{mathx}{m}{n}
\DeclareFontSubstitution{U}{mathx}{m}{n}
\DeclareMathAccent{\widebar}{0}{mathx}{"73}

\newcolumntype{d}[1]{D{.}{.}{#1}}
\newcolumntype{R}{@{\extracolsep{3cm}}r@{\extracolsep{0pt}}}

\DeclareMathOperator{\Sim}{sim}

\newcommand{\invariantNet}{{\bf F}_\text{ae}}
\newcommand{\variantNet}{{\bf F}_\text{ist}}
\newcommand{\variantMLP}{{\bf H}}

\newcommand{\affineTransform}{\bM_{t \rightarrow q}}

\newcommand{\maskQueryFeat}{{\bf m}_{\calQ_k}}
\newcommand{\maskTemplateFeat}{{\bf m}_{\calT_k}}

\newcommand{\ourmethod}{GigaPose\xspace}

\newcommand{\featuretemplate}{{\bf t}}
\newcommand{\featurequery}{{\bf q}}
\newcommand{\template}{\mathcal{T}}
\newcommand{\query}{\mathcal{Q}}

\newcommand{\Routplane}{\bR_\text{ae}}

\newcommand{\croppingQuery}{\bM_{\mathcal{Q}}}
\newcommand{\croppingTemplate}{\bM_{\mathcal{T}}}
\newcommand{\fullAffineTransform}{\bM_{\mathcal{T} \rightarrow \mathcal{Q}}}

\newcommand{\TODO}{TODO}

\def\cvprPaperID{3513} 
\def\confName{CVPR}
\def\confYear{2024}

\title{GigaPose: Fast and Robust Novel Object Pose Estimation via One Correspondence} 

\newcommand{\namesep}{\hspace{0.8em}}

\author{
Van Nguyen Nguyen$^{1}$\namesep
Thibault Groueix$^{2}$\namesep
Mathieu Salzmann$^{3}$\namesep
Vincent Lepetit$^{1}$\\
{$^{1}$LIGM, \'Ecole des Ponts}\namesep{$^{2}$Adobe}\namesep{$^{3}$EPFL}\\[0.2cm]
}

\begin{document}

\definecolor{darkgreen}{RGB}{0,110,0}
\definecolor{darkred}{RGB}{170,0,0}
\def\greencheckmark{\textcolor{darkgreen}{\checkmark}}
\def\redxmark{\textcolor{darkred}{\text{\ding{55}}}}  %

\addeditor{nguyen}{NG}{0.7, 0.0, 0.7}
\addeditor{thibault}{TG}{0.0, 0.0, 0.8}
\addeditor{vincent}{VL}{0.0, 0.5, 0.0}
\addeditor{mathieu}{MS}{0.1, 0.5, 0.9}
\showeditstrue
\showeditsfalse

\newcommand\customparagraph[1]{\vspace{0.7em}\noindent\textbf{#1}}

\twocolumn[\maketitle\vspace{-5em}\newlength{\plotheight}
\setlength\plotheight{2.55cm}
\setlength\lineskip{1.pt}
\setlength\tabcolsep{3.5pt} 

\vspace{0.5cm}

\begin{center}
{\small
\begin{tabular}{
>{\centering\arraybackslash}m{\plotheight}
>{\centering\arraybackslash}m{\plotheight}
>{\centering\arraybackslash}m{\plotheight}
c
>{\centering\arraybackslash}m{\plotheight}
>{\centering\arraybackslash}m{\plotheight}
>{\centering\arraybackslash}m{\plotheight}
}
Ground-truth \& & MegaPose~\cite{megapose} & \ourmethod & &
Reconstruction \& & MegaPose~\cite{megapose} & \ourmethod \\
Input segmentation & {\footnotesize (1.68 s / detection)} & {\footnotesize (0.048 s / detection)} & &
 Input segmentation & {\footnotesize (1.68 s / detection)} & {\footnotesize (0.048 s / detection)}\\
\frame{\includegraphics[height=\plotheight,]{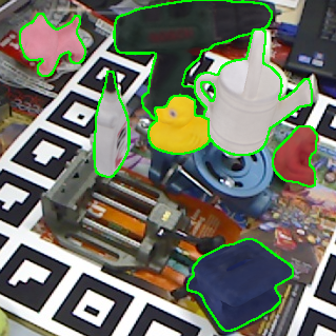}}&
\frame{\includegraphics[height=\plotheight, ]{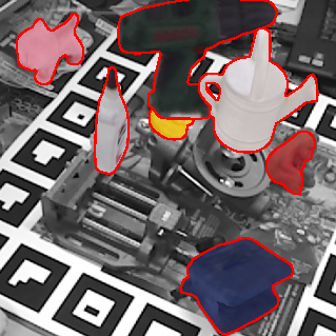}} &
\frame{\includegraphics[height=\plotheight, ]{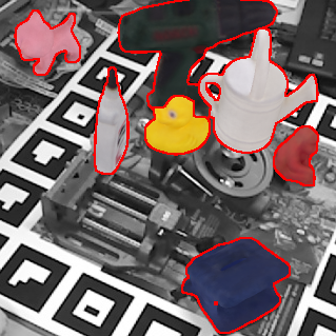}} & &
\frame{\includegraphics[height=\plotheight,]{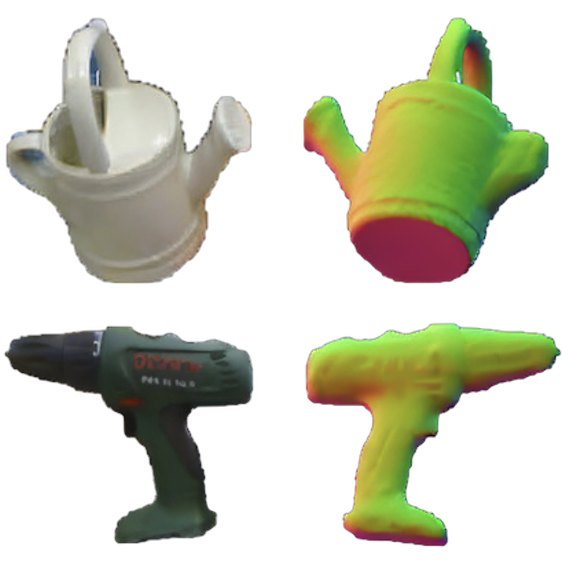}} &
\frame{\includegraphics[height=\plotheight, ]{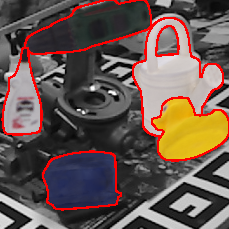}} &
\frame{\includegraphics[height=\plotheight, ]{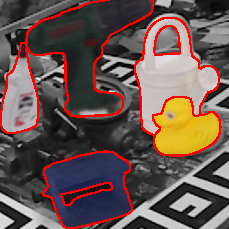}}\\ 

\frame{\includegraphics[height=\plotheight, ]{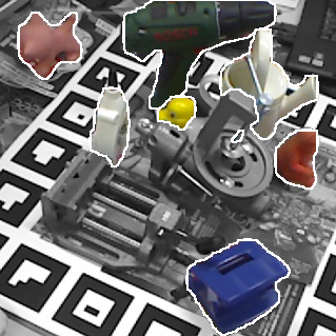}} &
\frame{\includegraphics[height=\plotheight, ]{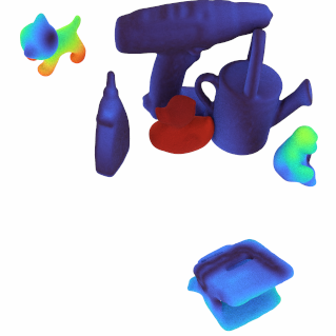}} &
\frame{\includegraphics[height=\plotheight, ]{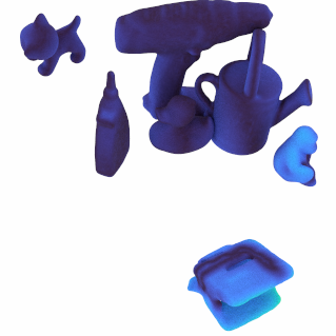}} & &
\frame{\includegraphics[height=\plotheight, ]{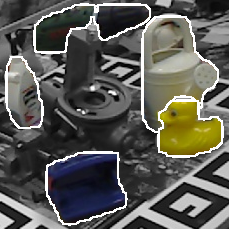}} &
\frame{\includegraphics[height=\plotheight, ]{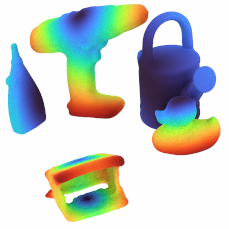}} &
\frame{\includegraphics[height=\plotheight, ]{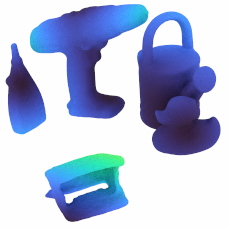}}\\ 

\begin{tabular}{
cc
}
 \hspace{1.9cm} Using ground-truth 3D models & \hspace{3.2cm} 
 Using 3D models predicted from a single image \\
\end{tabular}
\end{tabular}
}
    \vspace{-3pt}
    \captionof{figure}{ 
    {\bf Comparison of our method \ourmethod with MegaPose~\cite{megapose}.} \ourmethod is (i) more robust to noisy segmentation, often due to occlusions, (ii) more accurate with 3.5 \% average precision improvement on the BOP benchmark~\cite{sundermeyer2023bop}, and (iii) significantly faster with a speed up factor of 35$\times$ per detection for coarse object pose estimation stage (0.048 s vs 1.68 s). Left example compares the results using accurate 3D models, while the right example shows the results with 3D models predicted from a single image by Wonder3D~\cite{long2023wonder3d}. The bottom row shows the input segmentation, and the depth error heatmap of each detected object with respect to the ground truth pose, i.e the distance between each 3D point in the ground-truth depth map and its position with the predicted pose (legend: 0~cm \includegraphics[height=1.8mm]{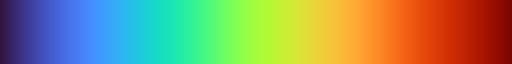} 10~cm).}
    \label{fig:teaserTask}
\end{center}

\vspace{0.5cm}\bigbreak]

\vspace*{-25pt}
\begin{abstract}

We present \ourmethod, a fast, robust, and accurate method for CAD-based  novel object pose estimation in RGB images. \ourmethod first leverages discriminative ``templates'', rendered images of the CAD models, to recover the out-of-plane rotation and then uses patch correspondences to estimate the four remaining parameters. Our approach samples templates in only a two-degrees-of-freedom space instead of the usual three and matches the input image to the templates using fast nearest-neighbor search in feature space, results in a speedup factor of 35$\times$ compared to the state of the art.  Moreover, \ourmethod is significantly more robust to segmentation errors. Our extensive evaluation on the seven core datasets of the BOP challenge demonstrates that it  achieves state-of-the-art accuracy and can be seamlessly integrated with existing refinement methods. Additionally, we show the potential of \ourmethod with 3D models predicted by recent work on 3D reconstruction from a single image, relaxing the need for CAD models and making 6D pose object estimation much more convenient. Our source code and trained models are publicly available at \href{https://github.com/nv-nguyen/gigaPose}{https://github.com/nv-nguyen/gigaPose}.
\end{abstract}
\thispagestyle{plain}
\pagestyle{plain}
\vspace*{-30pt}

\section{Introduction}

\label{sec:introduction}
\vspace*{-5pt}

6D object pose estimation has significantly improved over the past decade~\cite{kehl-iccv17-ssd6d,rad-iccv17-bb8,tekin-cvpr18-realtimeseamlesssingleshot,li-eccv18-deepim,zakharov-iccv19-dpod, labbe-eccv20-cosypose, Wang_2021_GDRN, liu2022gdrnpp_bop}. However, supervised deep learning methods, despite remarkable accuracy, are cumbersome to deploy to an industrial setting. Indeed, for each novel object, the pose estimation model needs to be retrained using newly-acquired data, which is impractical: Retraining typically takes several hours or days~\cite{labbe-eccv20-cosypose,liu2022gdrnpp_bop}, and the end users might not have the skills to retrain the model.

To fulfill the needs of such industrial settings, CAD-based novel object pose estimation, which focuses on estimating the 6D pose of \textit{novel} objects (i.e., objects only available at inference time, not during training), has garnered attention and was introduced in the latest BOP challenge~\cite{hodan2023bop}. Current approaches involve three main steps: object detection and segmentation, coarse pose estimation, and refinement. While object detection and segmentation has been recently addressed by CNOS~\cite{nguyen2023cnos},  refinement has been also addressed effectively with render-and-compare approaches~\cite{megapose,tremblay2023diff}. However, existing solutions to coarse pose estimation still suffer from low inference speed and sensitivity to segmentation errors. We thus focus on this step in this paper.

The low inference speed stems from how existing coarse pose estimation methods rely on templates~\cite{nguyen2022templates,chen_fusion,megapose}. Among them, MegaPose~\cite{megapose} has been widely adopted and integrated into various pipelines, notably the BOP challenge-winner GenFlow~\cite{genflow}~\footnote{GenFlow's description and performance are available in the BOP challenge~\cite{hodan2023bop} but it remains unpublished at the time of writing this paper.}. However, the complexity of MegaPose is linear in the number of templates, as it matches the input images against the templates by running a network on each image-template pair. As a result,  the methods based on MegaPose require more than 1.6 seconds \textit{per detection}.

Sensitivity to detection and segmentation errors, often due to occlusions, is a common issue for template-based approaches~\cite{nguyen2022templates,megapose}. As illustrated in Figure~\ref{fig:teaserTask}, the segmentation of occluded objects such as the ``duck''~(left example), results in a scale and translation mismatch when cropping the test image and templates. Additionally, the erroneous segments may include noisy signal from the other objects or the background, which results in numerous outlier matches between the input image and the templates. 

To address these two major limitations, we introduce \ourmethod, a novel approach for CAD-based coarse object pose estimation. \ourmethod makes several technical contributions towards speed and robustness and can be seamlessly integrated with any refinement method for CAD-based novel object pose estimation to achieve state-of-the-art accuracy.

The key idea in \ourmethod is to find the right trade-off between the use of templates, which have been shown to be extremely useful for estimating the pose of novel objects, and patch correspondences, which lead to better robustness and more accurate pose estimates. More precisely, we propose to rely on templates to estimate two degrees of freedom~(DoFs)---azimuth and elevation---as varying these angles changes the appearance of an object in complex ways, which templates excel at capturing effectively.  Our templates are represented with local features that are trained to be robust to scaling and in-plane rotations. Matching the input image with the templates based on these local features yields robustness to segmentation errors.

To estimate the remaining 4 DoFs---in-plane rotation and 3D translation decomposed into 2D translation and 2D scale, we rely on patch correspondences between the input image and the template candidates. Given a template candidate, we match its local features with those of the input image, which gives us 2D-2D point correspondences. Instead of simply exploiting the matched point coordinates and use a P$n$P algorithm~\cite{moreno2007accurate} to estimate the pose as done in previous works~\cite{rad-iccv17-bb8,hu-cvpr20-singlestage6dobjectposeestimation,hu-cvpr19-segmentationdriven6dobjectposeestimation}, we also exploit their appearances: We show that it is possible to predict the in-plane rotation  and relative scale between the input image and the template from local features computed at the matched points. The remaining 2D translation is obtained from the positions of these matched points, allowing the estimation of the four DoFs from a single correspondence. To robustify this estimate, we combine this process with RANSAC.

We experimentally demonstrate that our balance between the use of templates and patch correspondences effectively addresses the two issues in coarse pose estimation. Indeed, our method relies on a sublinear nearest-neighbor template search, successfully addressing the low inference speed issue with a speedup factor of 35$\times$ per detection compared to to MegaPose~\cite{megapose}.  Furthermore, the two steps of our method are particularly robust to segmentation errors.

We also demonstrate that \ourmethod can exploit a 3D model reconstructed from a single image by a diffusion-based model~\cite{long2023wonder3d,liu2023syncdreamer,liu2023zero,qian2023magic123,poole2022dreamfusion,sjc} instead of an accurate CAD model. Despite the inaccuracies of the predicted 3D models, our method can recover an accurate 6D pose as shown on Figure~\ref{fig:teaserTask}. This relaxes the need for CAD models and makes 6D pose object detection much more convenient.

In summary, our contribution is a novel RGB-based method for CAD-based novel object coarse pose estimation from a single correspondence that is significantly faster, more robust, and more accurate than existing methods. We demonstrate this through extensive experiments on the seven core datasets of the BOP challenge~\cite{sundermeyer2023bop}.

\vspace*{-3pt}
\section{Related Work}

\label{sec:relatedwork}
\vspace*{-8pt}
\customparagraph{Seen object pose estimation.}  Early works on 6D pose estimation have introduced diverse benchmarks to evaluate the performance of their approaches~\cite{brachmann-eccv14-learning6dobjectposeestimation,hodan-wacv17-tless,Hodan_undated-sl,doumanoglou2016recovering,drost2017introducing,kaskman2019homebreweddb,Xiang2018-dv, sundermeyer2023bop}. This data and its ground truth have powered many deep learning-based methods~\cite{kehl-iccv17-ssd6d,rad-iccv17-bb8, li-eccv18-deepim,tekin-cvpr18-realtimeseamlesssingleshot,li-iccv19-cdpn,zakharov-iccv19-dpod,park-iccv19-pix2pose,hu-cvpr20-singlestage6dobjectposeestimation,labbe-eccv20-cosypose,liu2022gdrnpp_bop}. Some of them show remarkable performance in terms of run-time and accuracy~\cite{labbe-eccv20-cosypose,liu2022gdrnpp_bop}. However, these approaches  require long and expensive training, such as the state-of-the-art methods~\cite{labbe-eccv20-cosypose, liu2022gdrnpp_bop} require several hours for training for a single object, making them too cumbersome for many practical applications in robotics and AR/VR.


To avoid the need for re-training when dealing with new object instances, one approach is to train on object categories by assuming that the testing objects belong to a known category~\cite{lin2022icra, Wang_2019_NOCS, chen2020category,li2022polarmesh,manhardt2020cps++,manuelli2019kpam,irshad2022shapo,irshad2022centersnap}. These category-level pose estimation methods, however, cannot generalize to objects beyond the scope of the training categories.  By contrast, our method operates independently of any category-level information and seamlessly generalizes to novel categories.

\definecolor{darkgreen}{RGB}{0, 100, 0}
\definecolor{lighterblue}{RGB}{42, 160, 255}
\begin{figure*}[!t]
    \begin{center}
    \includegraphics[width=1\linewidth]{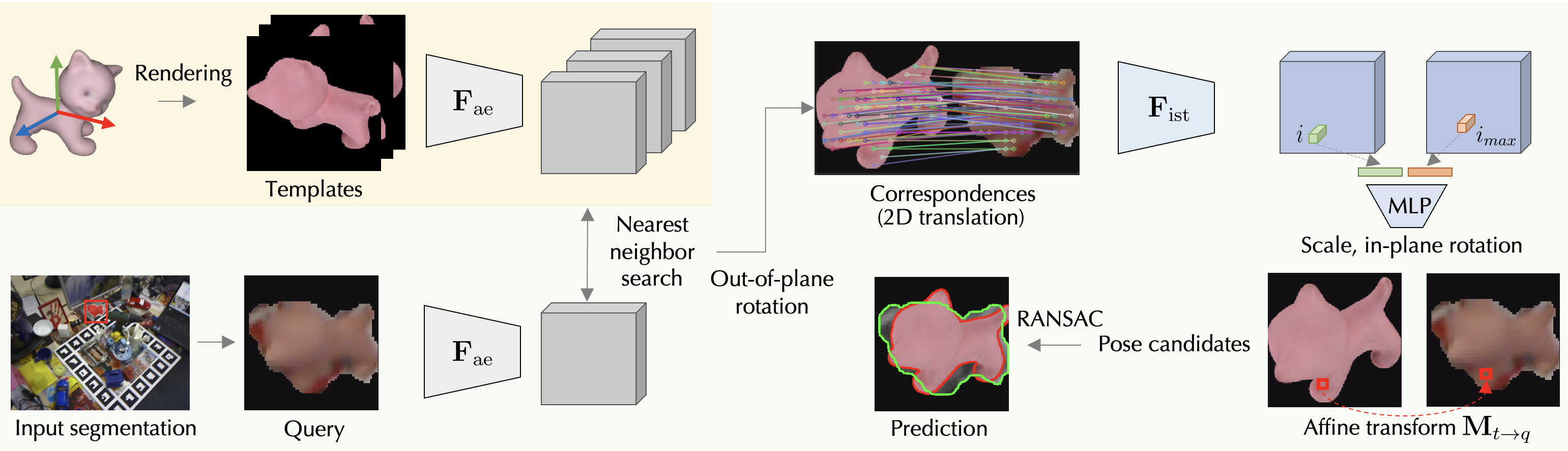}
    \end{center}
    \vspace{-12pt}
    \caption{
    {\bf Overview.} We first onboard each novel object by rendering 162 templates, spanning the spectrum of out-of-plane rotations. We also extract dense features using $\invariantNet$ from each of the templates. At runtime, given a query image segmented with CNOS~\cite{nguyen2023cnos}, we process it (by masking the background, cropping on the segment, adding padding then resizing), and extracting features with $\invariantNet$. We retrieve the nearest template to the segment using the similarity metric detailed in Section~\ref{sec:azimuthElevation}. Further, 2D scale and in-plane rotation are computed from a single 2D-2D correspondence using $\variantNet$ and two lightweight MLPs. The 2D position of the correspondences also gives us the 2D translation which is used with 2D scale, in-plane rotation to create the affine transformation $\affineTransform$, mapping the nearest template to the query image. This enables us to recover the complete 6D object pose from a single correspondence. Finally, we use RANSAC to robustly find the best pose candidate. Onboarding takes {11.5} seconds per object and inference takes {48} milliseconds per detection on average. }
\label{fig:inferenceFramework}
\end{figure*}
\vspace*{-2pt}
\customparagraph{Novel object pose estimation.}
Several techniques have been explored to improve the generalization of object pose estimation methods~\cite{Pitteri20203DOD,ausserlechner2023zs6d,okorn2021zephyr,nguyen2022templates,chen_fusion,shugurov_osop_2022,tremblay2023diff,megapose,nguyen_pizza_2022,nguyen2024nope,ornek2023foundpose,lin2023sam}.  These can be roughly divided into feature-matching methods \cite{Pitteri20203DOD,ausserlechner2023zs6d} and template matching ones~\cite{okorn2021zephyr,nguyen2022templates,chen_fusion,shugurov_osop_2022,megapose,ausserlechner2023zs6d,nguyen2024nope,ornek2023foundpose}. Feature-matching methods extract local features from the image, match them to the given 3D model and then use a variant of the P$n$P algorithm~\cite{moreno2007accurate} to recover the 6D pose from the 3D-to-2D correspondences.  By contrast, template matching methods first render synthetic templates of the CAD models, and then use a deep network to compute a score for each input image-template pair, thus aiming to find the template with the most similar pose to the input image.

At the last BOP challenge~\cite{hodan2023bop},  CAD-based \textit{novel} object pose estimation was introduced as a new task, using CNOS~\cite{nguyen2023cnos} as the default detection method. MegaPose~\cite{megapose} and ZS6D~\cite{ausserlechner2023zs6d} showed promising results for this new task. Nevertheless, MegaPose's run-time was highlighted as a significant limitation due to the need for a forward pass through a coarse pose estimator to compute a classification score for every (query, template) comparison. ZS6D relies on DINOv2 features~\cite{oquab2023dinov2} and the similarity metric of \cite{nguyen2022templates} to predict sparse 3D-to-2D correspondences. Unfortunately, their experiments do not evaluate the model's sensitivity to segmentation errors. In contrast, we present extensive evaluations  and demonstrate that GigaPose outperforms  both MegaPose and ZS6D. Our method can also be seamlessly integrated with any refinement method.

\vspace*{-2pt}
\customparagraph{Correspondence-based object pose estimation.}
A classical approach to solving the 6D object pose estimation problem is to establish 3D-to-2D correspondences and compute the pose with a P$n$P algorithm~\cite{rad-iccv17-bb8,tekin-cvpr18-realtimeseamlesssingleshot,hu-cvpr19-segmentationdriven6dobjectposeestimation,hu-cvpr20-singlestage6dobjectposeestimation,zakharov-iccv19-dpod,peng-cvpr19-pvnet,li-iccv19-cdpn}, which requires at least four 3D-to-2D point correspondences.  Because we first match the input image against templates and estimate the four remaining DoFs from a single 2D-to-2D match, we only need one correspondence to determine the 6D object pose. Our experimental evaluation shows that our method outperforms ZS6D, which, as stated above, is a state-of-the-art method relying on 3D-to-2D correspondences.

\vspace*{-5pt}
\section{Method}
\label{sec:method}
\label{sec:overview}

\begin{figure*}[!t]
    \begin{center}
    \includegraphics[width=0.95\linewidth]{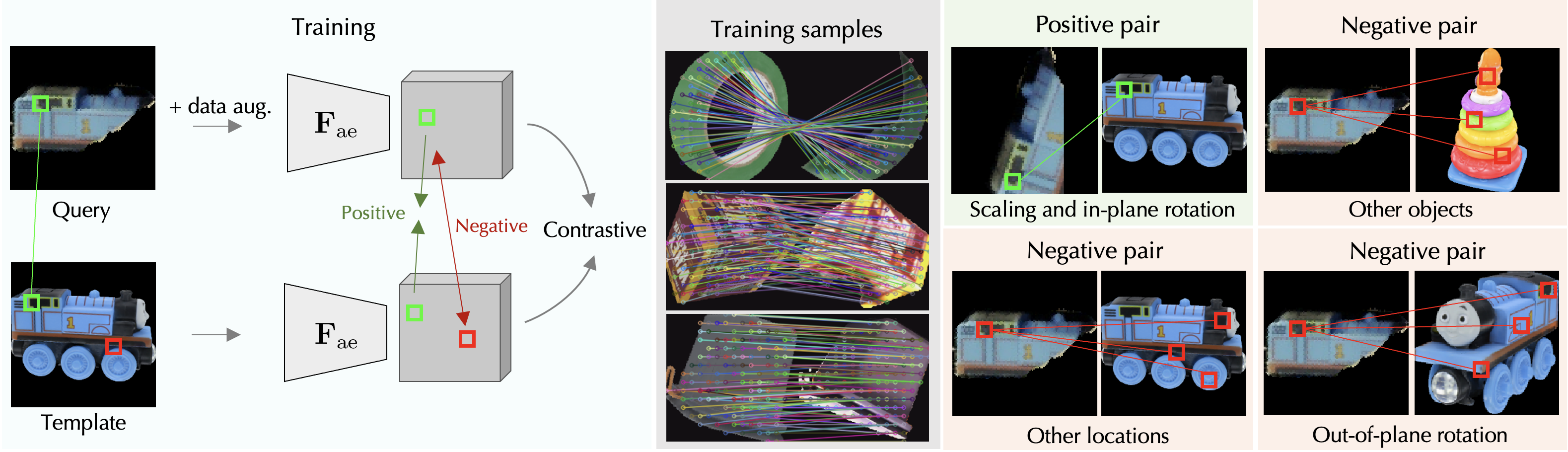}
    \end{center}
    \vspace{-12pt}
    \caption{
        \label{fig:trainingFramework}
        {\bf Contrastive training of $\invariantNet$.} We use pairs made of a query image and a template to train a network using local contrastive learning as detailed in Section~\ref{sec:azimuthElevation}. Middle: Training samples provided by \cite{megapose}, and the 2D-2D correspondences created from ground-truth 3D information used to generate positive and negative pairs. Right: We seek local features that vary with the out-of-plane rotation, but are invariant to in-plane rotation and scaling. Thus, positive pairs are made of corresponding patches under scaling and in-plane rotation changes, and negative pairs are made of corresponding patches under different out-of-plane rotations, patches that do not correspond, or that come from  different objects. 
        }
\end{figure*}

Figure~\ref{fig:trainingFramework} provides an overview of \ourmethod.
Given the 3D model of an object of interest, we render templates and extract their dense features using a Vision-Transformer~(ViT) model $\invariantNet$. 
Then, given an input image, we detect the object of interest and segment it using an off-the-shelf method CNOS~\cite{nguyen2023cnos}. \ourmethod extracts dense features from the input image at the object location using $\invariantNet$ again.  We select the template most similar to the input image using a similarity metric based on the dense features, detailed in Section~\ref{sec:azimuthElevation}.
This gives us the azimuth and elevation of the camera. 


To estimate the remaining DoFs, we look for corresponding patches between the input image and its most similar template. From one such pair of patches, we can directly predict two additional DoFs: the 2D scale $s$ and the in-plane rotation $\alpha$, by feeding two lightweight MLPs the features for the two patches extracted by another feature extractor denoted as $\variantNet$. Note that the features extracted by $\invariantNet$ are not suitable here, as they discard information about scale and in-plane rotation by design. The image locations of the  corresponding patches also give us directly the last two DoFs: the 2D translation $(\bt_x, \bt_y)$. From the scale and 2D translation, we can estimate the 3D translation.  We use a RANSAC framework and iterate over different pairs of patches to find the optimal pose. We detail the training of $\variantNet$ and the MLPs, and the RANSAC scheme in Section~\ref{sec:onecorres}. 

\subsection{Generating Templates}
\vspace*{-3pt}

In contrast to other approaches~\cite{nguyen2022templates,chen_fusion}, we do not generate templates for both in-plane and out-of-plane rotations, as this yields thousands of templates. Instead, we decouple the 6~DoFs object pose into out-of-plane rotation, in-plane rotation, and 3D translation (2D translation and scale). Given the out-of-plane rotation, finding the scale and in-place rotation is indeed a 2D problem only. We thus create much less templates and push the estimation of the other DoFs to a later stage in the pipeline (see Section~\ref{sec:onecorres}). 

In practice, we use 162 templates. These are generated from viewpoints defined in a regular icosphere which is created by subdividing each triangle of the icosphere primitive of Blender into four smaller triangles. This has been shown in previous works~\cite{nguyen2023cnos,ausserlechner2023zs6d,nguyen2022templates} to provide well-distributed view coverage of CAD models.


\vspace*{-2pt}

\subsection{Predicting Azimuth and Elevation}
\vspace*{-3pt}
\label{sec:azimuthElevation}

\paragraph{Training the feature extractor $\invariantNet$.}
$\invariantNet$ extracts dense features from both the input image and each of the templates \textit{independently}. Compared to estimating features jointly, this approach eliminates the need for extensive feature extraction at runtime, a process that scales linearly with the number of templates and in-plane rotations considered. Instead, we can offload the computation of the features for each template to an onboarding stage for each novel object. We now describe how we train the feature extractor $\invariantNet$ and how we design the similarity metric to compare the template and query features.

The extracted features aim to match a query image to a set of templates with different out-of-plane rotations, but with fixed scale, in-plane rotation, and translation. The features should thus be invariant to scale, in-plane rotation, and 2D translation, but be sensitive to out-of-plane rotation.

We achieve this with a local contrastive learning scheme. The main difficulty lies in defining the positive and negative patch pairs. Figure~\ref{fig:trainingFramework} illustrates our training procedure. 
We construct batches of $B$ image pairs $(\query_k, \template_k)$, such that the query $\query_k$ is a rendering of a 3D object in any pose, and the template $\template_k$ is another rendering of that object with the same out-of-plane rotation but different in-plane rotation, scale, and 2D translation. Because we have access to the 3D model, we can compute ground-truth 2D-to-2D correspondences to create positive and negative pairs. We detail in the supplementary material how we compute these correspondences. 


Additionally, since our goal is to close the domain gap between real images and synthetic renderings, we apply color augmentation along with random cropping and in-plane rotation to the input pairs. We use the training sets provided by the BOP challenge~\cite{sundermeyer2023bop}, originally sourced from MegaPose~\cite{megapose}. These datasets consist of 2 million images and are generated from CAD models of Google Scanned Objects~\cite{downs2022google} and ShapeNet~\cite{chang2015shapenet}  using BlenderProc~\cite{denninger2019blenderproc}. We show typical training samples in the middle part of Figure~\ref{fig:trainingFramework}.

We pass each image  $\query_k$ and $\template_k$ independently through $\invariantNet$ to extract dense feature maps $\featurequery_k$ and $\featuretemplate_k$. Below, we use the superscript $i$ to denote a 2D location in the local feature map. Note that because of the downsizing done by the ViT, each location $i$ in the feature grid corresponds to a 14$\times$14 patch in the respective input image. Each feature map has a respective segmentation mask $\maskQueryFeat$ and $\maskTemplateFeat$ corresponding to the foreground of the object in the images $\query_k$ and $\template_k$.

For a location $i$ in the query feature map $\featurequery_k$, we denote by $i^*$ the corresponding location in the template feature map. We arrange the query patches $(\featurequery_k^{i})$ and their corresponding patch $(\featuretemplate_k^{i^*})$ in a square matrix such that the diagonal contains the positive pairs, and all other entries serve as negative pairs. For each pair $(\query_k, \template_k)$, we thus obtain $|\maskQueryFeat|$ positive pairs and $|\maskQueryFeat| \times (|\maskQueryFeat|-1)$ negative pairs.

To improve the efficiency of contrastive learning, we use additional negative pairs from a query image $\query_k$ and a template $\template_{k^{\prime}}$, where $k^{\prime}\neq k$ in the current batch. This process results in total in $\left(\sum_{k=1}^B |\maskQueryFeat|\right)^2 - \sum_{k=1}^B |\maskQueryFeat|$  negative pairs for the current batch. We train $\invariantNet$ to align the representations of the positive pairs while separating the negative pairs using the InfoNCE loss~\cite{Oord2018infoNCE}:
\begin{equation}\label{loss:infoNCE}
    \calL_{\text{out}} = - \sum_{k=1}^B \sum_{i=1}^{|\maskQueryFeat|} \ln \frac{e^{S(\featurequery_k^i, \featuretemplate_k^{i^*})/\tau}}
    {\sum_{(k^{\prime}, i^{\prime}) \neq (k, i^*)}  e^{S(\bq_{k}^{i}, {\bf t}_{k^{\prime}}^{i^{\prime}})/\tau}} \> ,
\end{equation}
%
%
%
%
where $S(., .)$ is the cosine similarity between the local image features computed by the network $\invariantNet$. The temperature parameter $\tau$ is set to $0.1$ in our experiments. 

Since the positive pairs of patches have different scales and in-plane rotations, our network $\invariantNet$ learns to become invariant to these two factors, as demonstrated in our experiments. We initialize our feature extractor $\invariantNet$ as  DINOv2~\cite{oquab2023dinov2} pretrained on ImageNet, because it has proven to be highly effective in extracting features for vision tasks. 
\vspace*{-10pt}
\paragraph{Azimuth and elevation prediction.}  We define a pairwise similarity metric for each query-template $(\query, \template)$ pair  with their respective dense feature grid $(\featurequery, \featuretemplate)$ and feature segmentation masks $(\text{m}_\calQ, \text{m}_\calT)$.

For each local query feature $\featurequery^i$, corresponding to patch at location $i$, we compute its nearest neighbor in the template features $\featuretemplate$, denoted as $\featuretemplate^{i_{\text{max}}}$, as
\begin{equation}
i_{\text{max}} = \argmax_{{j | \text{m}_{\mathcal{T}}^{j}>0}} S\left(\featurequery^{i},\featuretemplate^{j}\right) \> .
\label{eq:nearest}
\end{equation}
This nearest neighbor search yields a list of correspondences $\{ ({i, i_{\text{max}}}) \}$. To improve the robustness of our method against outliers, we keep only the correspondences $\{ ({i, i_{\text{max}}}) \}$ having a similarity score $\geq$ 0.5. The final similarity for this ($\query$, $\template$) pair is defined as the mean of all the remaining correspondences, weighted by their similarity score:
\begin{equation}
    \Sim(\featurequery, \featuretemplate) = \frac{1}{| \text{m}_{\mathcal{Q}}|} \sum_i  \text{m}_{\mathcal{Q}}^{i}  S \left(\featurequery^{i}, \featuretemplate^{i_{\text{max}}} \right) \>.
\label{eq:sim}
\end{equation}
We compute this score for all templates $\template_{k}$ ($1 \leq k \leq 162$) and find the top-$K$ candidates yielding the most similar out-of-plane rotations. This nearest neighbor search is very fast and delivers results within {\TODO} milliseconds. In practice, we experiment with $K=1$ and $K=5$. For the latter, the final template is selected by the RANSAC-based estimation detailed in Section~\ref{sec:onecorres} below. 



\subsection{Predicting the Remaining DoFs}
\label{sec:onecorres}
Once we have identified the template candidates, we seek to estimate the remaining 4~DoFs, i.e., in-plane rotation, scale, and 2D translation, which yield the affine transformation $\affineTransform$ transforming each template candidate $\template$ to the query image $\query$. Specifically, we have
\begin{equation}
\begin{aligned}
\affineTransform &= \begin{bmatrix}
    s \cos(\alpha) & -s \sin(\alpha) & \bt_x \\
    s \sin(\alpha) & s \cos(\alpha) & \bt_y \\
    0 & 0 & 1 \\
\end{bmatrix} \> ,
\end{aligned}
\label{eq:affine}
\end{equation}
where $s$ is the 2D scaling factor, $\alpha$ is the relative in-plane rotation, and $\left[\mathbf{t}_x, \mathbf{t}_y\right]$ is the 2D translation between the input query image $\query$ and the template $\template$.

\paragraph{Training the feature extractor $\variantNet$ and the MLP.}
We have already obtained from the features of $\invariantNet$ a list of 2D-2D correspondences $\{ ({i, i_{\text{max}}}) \}$. Each correspondence can inherently provide 2D translation $\left[\mathbf{t}_x, \mathbf{t}_y\right]$ information through the patch locations $i$ and $i_{\text{max}}$. To recover the remaining 2 DoFs, scale $\mathbf{s}$ and in-plane rotation $\alpha$, we train deep networks to directly regress these values from a single 2D-2D correspondence. Since the feature extractor $\invariantNet$ is invariant to in-plane rotation and scaling, the corresponding features cannot be used to regress those values, hence we have to train another feature extractor we call $\variantNet$. Given a 2D-2D match from a pair $(\query, \template)$, and their corresponding feature computed by $\variantNet$, we pass them through two small MLPs, which outputs directly $\alpha$ and $\mathbf{s}$. This enables us to predict 2D scale and in-plane rotations for each 2D-2D correspondence. To address the $2\pi$ periodicity of in-plane rotation, we predict $\left[\cos(\alpha_k^l), \sin(\alpha_k^l)\right]$ instead of $\alpha_k^l$.

We train jointly both $\variantNet$ and the MLPs on the same data samples as $\invariantNet$ using the loss:
%
\begin{equation}\label{loss:viariantnet}
    \calL_{\text{inp}} =  \sum_{k=1}^B \sum_{i=1}^{n_k} \left[ \left(\ln(s_{k}^{i}) - \ln(s_{k}^{*})\right)^2 + \text{geo}(\alpha_{k}^i, \alpha_{k}^{*})\right],
\end{equation}
where $s_{k}^{*}$ and $\alpha_{k}^{*}$ are the ground-truth scale and in-plane rotation  between $\query$ and $\template_{k}$, and $\text{geo}(\cdot,\cdot)$ indicates the geodesic loss defined as
{\small
\begin{equation}\label{loss:geodesic}
\begin{aligned}
    \text{geo}(\alpha_1, \alpha_2)=\text{acos}\bigl(\text{cos}(\alpha_1)\text{cos}(\alpha_2) + 
    \text{sin}(\alpha_1)\text{sin}(\alpha_2)\bigr) \> .
\end{aligned}
\end{equation}
}
\paragraph{RANSAC-based $\affineTransform$ estimation.}
For each template $\template$, we employ RANSAC on each $\affineTransform$ predicted by each correspondence and validate them against the remaining correspondences using a 2D error threshold of $\delta$. In practice, we set $\delta$ to the size of a patch, corresponding to an error of 14 pixels in image space. The final prediction for $\affineTransform$ is determined by the correspondence with the highest number of inliers. The complete 6D object pose can finally be recovered from the out-of-plane rotation, in-plane rotation, 2D scale and 2D translation using the explicit formula provided in the supplementary material.

We initialize $\variantNet$ with a modified version of ResNet18 \cite{he-cvpr16-deepresiduallearning} instead of the DINOv2 \cite{oquab2023dinov2} as DINOv2 is trained with random augmentations that includes in-plane rotations and cropping, making its features invariant to scale and in-plane rotation. Similarly to the features from $\invariantNet$,  we offload the feature computation of $\variantNet$ to the onboarding stage for all templates to avoid the computational burden at runtime. 


\newcommand{\omitdetection}[1]{}
\definecolor{navyblue}{RGB}{191, 209, 229} 
\definecolor{light_yellow}{RGB}{255,243,194}
\definecolor{orange}{RGB}{255,200,100}
\definecolor{light_red}{RGB}{255,150,150}
\begin{table*}[!t]
  \centering
  \small %
 \setlength{\tabcolsep}{2.5pt}
\begin{adjustbox}{max width=\textwidth}

\begin{tabular}{rlllcccccccc|c|c}
 \toprule  
 & \multirow{3}{*}{Method} & \multirow{3}{*}{Detections} & \multirow{3}{*}{Refinement} & & \textsc{lm-o} & \textsc{t-less} & \textsc{tud-l} & \textsc{ic-bin} & \textsc{itodd} & \textsc{hb} & \textsc{ycb-v} & \multirow{3}{*}{\footnotesize{\textsc{Mean}}}  & \multirow{3}{*}{\footnotesize{\textsc{Run-time}}}  \\ 
 
 \cmidrule(lr){6-12}
 &  &  &  \;\;\;\;\;\;\;\;\;\;\;\;\;\;\;\;\;\; \;\;\;\;\;\;\; \footnotesize{Num. instances}: &  & \footnotesize{\textsc{1445}} & \footnotesize{\textsc{6423}} & \footnotesize{\textsc{600}}  & \footnotesize{\textsc{1786}}  & 
  \footnotesize{\textsc{3041}}  & \footnotesize{\textsc{1630}}  & 
\footnotesize{\textsc{4123}}\\

 \midrule %

{\color{teal}\scriptsize 1} & OSOP \cite{shugurov_osop_2022} & OSOP \cite{shugurov_osop_2022} & -- & & 27.4 & 40.3 & -- & --  & -- & -- & 29.6 & --  & -- \\ %

{\color{teal}\scriptsize 2} & MegaPose \cite{megapose} & Mask R-CNN \cite{he2017mask} & -- & & 18.7 & 19.7 & 20.5 & 15.3 & 8.00 & 18.6 & 13.9 & 16.2 & -- \\ %

\cmidrule(lr){1-13}

{\color{teal}\scriptsize 3} & ZS6D \cite{ausserlechner2023zs6d} & CNOS\cite{nguyen2023cnos} & -- & & \cellcolor{navyblue} 29.8 & 21.0 & -- & --  & -- & -- & \cellcolor{navyblue} 32.4 & --  & -- \\ %

{\color{teal}\scriptsize 4} & MegaPose \cite{megapose} & CNOS \cite{nguyen2023cnos} & -- & & 22.9 & 17.7 & 25.8 & 15.2 & 10.8 & 25.1 & 28.1 & 20.8 & 15.5 s \\ %

{\color{teal}\scriptsize 5} & GigaPose (Ours) & CNOS \cite{nguyen2023cnos} &  -- & & 29.6 & \cellcolor{navyblue} 26.4 & \cellcolor{navyblue} 30.0 & \cellcolor{navyblue} 22.3 & \cellcolor{navyblue} 17.5  & \cellcolor{navyblue} 34.1  &  27.8 & \cellcolor{navyblue} 26.8  & $\phantom{0}$0.4 s \\ %

 \midrule %

{\color{teal}\scriptsize 6} & MegaPose \cite{megapose}  & CNOS \cite{nguyen2023cnos} & MegaPose \cite{megapose} & &  49.9 & 47.7 &  \cellcolor{light_yellow}  65.3 & 36.7 & 31.5 & 65.4 & 60.1 & 50.9 & 17.0 s \\ %

{\color{teal}\scriptsize 7} & GigaPose (Ours)  & CNOS \cite{nguyen2023cnos} & MegaPose \cite{megapose} & &  \cellcolor{light_yellow} 55.7 & \cellcolor{light_yellow} 54.1 &  58.0 &  \cellcolor{light_yellow} 45.0 & \cellcolor{light_yellow} 37.6 & \cellcolor{light_yellow} 69.3 &  \cellcolor{light_yellow} 63.2 & \cellcolor{light_yellow} 54.7  &  $\phantom{0}$2.3 s \\ 

 \midrule %
{\color{teal}\scriptsize 8} & MegaPose \cite{megapose}  & CNOS \cite{nguyen2023cnos} & MegaPose + 5 Hypotheses\cite{megapose} &  & 56.0 & 50.7 &  \cellcolor{orange} 68.4 & 41.4 & 33.8 & 70.4 & 62.1 & 54.7 & 21.9 s\\ 

{\color{teal}\scriptsize 9} & GigaPose (Ours) & CNOS \cite{nguyen2023cnos} & MegaPose + 5 Hypotheses \cite{megapose} & & \cellcolor{orange} 59.8 &  \cellcolor{orange} 56.5 &  63.1 &  \cellcolor{orange} 47.3 &  \cellcolor{orange} 39.7 &  \cellcolor{orange} 72.2 &  \cellcolor{orange} 66.1 & \cellcolor{orange} 57.8  &  $\phantom{0}$7.7 s \\

\midrule %
{\color{teal}\scriptsize 10} & MegaPose \cite{megapose} & CNOS \cite{nguyen2023cnos} & GenFlow  + 5 Hypotheses \cite{genflow} & &  56.3 & 52.3 &  \cellcolor{light_red} 68.4 & 45.3 & 39.5 &  73.9 & 63.3 & 57.0 & 20.8 s \\ %

{\color{teal}\scriptsize 11} & GigaPose (Ours) & CNOS \cite{nguyen2023cnos} & GenFlow  + 5 Hypotheses \cite{genflow} & & \cellcolor{light_red} 63.1 & \cellcolor{light_red} 58.2 &  66.4 & \cellcolor{light_red} 49.8 & \cellcolor{light_red} 45.3 &  \cellcolor{light_red} 75.6 & \cellcolor{light_red} 65.2 & \cellcolor{light_red} 60.5 &  10.6 s \\ %
  \bottomrule
  \end{tabular}
  
\end{adjustbox}
    \vspace{-7pt}
  \caption{{\bf Results on the BOP datasets.}  We report the AR score on each of the seven core datasets of the BOP challenge and the mean score across datasets.  The best results with CNOS's detections \cite{nguyen2023cnos} without refinement are highlighted in \colorbox{navyblue}{blue}, with \nguyen{MegaPose's refinement using 1 hypothesis in \colorbox{light_yellow}{yellow}, and using 5 hypotheses in \colorbox{orange}{orange}, and with GenFlow's refinement using 5 hypotheses in \colorbox{light_red}{red}. }}
  \label{tab:bop}
\end{table*}

\customparagraph{Implementation details.} 
We use the input image of size 224 $\times$224, resulting in features of size 16$\times$16$\times$1024 and 16$\times$16$\times$256 via the networks $\invariantNet$ and $\variantNet$ respectively. We train our networks using the Adam optimizer with an initial learning rate of 1e-5 for $\invariantNet$ and 1e-3 for $\variantNet$. The training process takes less than 10 hours when using four V100 GPUs. All the inference experiments are run on a single  V100 GPU. 

\vspace*{-5pt}
\section{Experiments}
\vspace*{-6pt}
In this section, we first describe our experimental setup~(Section~\ref{sec:exp_setup}). Next, we compare our method with previous works~\cite{megapose,ausserlechner2023zs6d,shugurov_osop_2022} on the seven core datasets of the BOP challenge~\cite{hodan2023bop}~(Section~\ref{sec:compare_SOTA}). We conduct this comparison to evaluate our method's accuracy, runtime performance, and robustness to segmentation errors, highlighting our contributions. Finally, we present an ablation study that explores different settings of our method~(Section~\ref{sec:ablation}).
\vspace*{-5pt}
\subsection{Experimental Setup}
\label{sec:exp_setup}
\vspace*{-3pt}
\paragraph{Evaluation Datasets.} We evaluate our method on the seven core datasets of the BOP challenge~\cite{hodan2023bop}: LineMod Occlusion~(LM-O)~\cite{brachmann-eccv14-learning6dobjectposeestimation}, T-LESS~\cite{hodan-wacv17-tless}, TUD-L~\cite{Hodan_undated-sl}, IC-BIN~\cite{doumanoglou2016recovering}, ITODD~\cite{drost2017introducing}, HomebrewedDB~(HB)~\cite{kaskman2019homebreweddb} and YCB-Video~(YCB-V)~\cite{Xiang2018-dv}. These datasets consist of a total of 132 different objects and 19048 testing instances, presented in cluttered scenes under partial occlusions. It is worth noting that, in contrast to the \textit{seen} object setting, the \textit{novel} object pose estimation setting is far from being saturated in terms of both accuracy and run-time.

\vspace*{-3pt}
\customparagraph{Evaluation metrics.} For all experiments, we use the standard BOP evaluation protocol~\cite{hodan-eccv20-bopchallenge2020on}, which relies on three metrics: Visible Surface Discrepancy~(VSD), Maximum Symmetry-Aware Surface Distance~(MSSD), and Maximum Symmetry-Aware Projection Distance~(MSPD). The final score, referred to as the average recall~(AR), is calculated by averaging the individual average recall scores of these three metrics across a range of error thresholds. 

\customparagraph{Baselines.} We compare our method with MegaPose~\cite{megapose}, ZS6D~\cite{ausserlechner2023zs6d}, and OSOP~\cite{shugurov_osop_2022}. 
As of the time of writing, the source codes for ZS6D and OSOP are not available. 
Therefore, we can only report their performance as provided in their papers, but not their run-time. 


\customparagraph{Refinement.}
To demonstrate the potential of \ourmethod, we have applied the refinement \nguyen{methods from MegaPose~\cite{megapose} and GenFlow~\cite{genflow} to our results. We extract the top-1 and the top-5 pose candidates and subsequently refine them using 5 iterations of MegaPose's refinement network~\cite{megapose} or GenFlow's refinement network~\cite{genflow}. For the top-5 hypotheses case, these refined hypotheses are scored by the coarse network of MegaPose~\cite{megapose}, and the best one is selected.}

\vspace*{-5pt}
\customparagraph{Pose estimation with a 3D model predicted from a single image.}
We use Wonder3D~\cite{long2023wonder3d} to predict a 3D model from a single image for objects from LM-O. We then evaluate the performance of MegaPose and our method using reconstructed models instead of the accurate CAD models provided by the dataset. Due to the sensitivity of Wonder3D to the quality of input images, we carefully select reference images.
More details about this setting are present in the supplementary material.
\vspace*{-5pt}
\newlength{\plotwidth}
\setlength\plotwidth{3.3cm}
\setlength\lineskip{2.5pt}
\setlength\tabcolsep{2pt} 

\begin{figure*}[!t]
\begin{center}
{\small
\begin{tabular}{
>{\centering\arraybackslash}m{\plotwidth}
>{\centering\arraybackslash}m{\plotwidth}
>{\centering\arraybackslash}m{\plotwidth}
>{\centering\arraybackslash}m{\plotwidth}
>{\centering\arraybackslash}m{\plotwidth}
>{\centering\arraybackslash}m{\plotwidth}
}
& \multicolumn{2}{c}{\textbf{coarse pose estimation}} &\multicolumn{2}{c}{\textbf{after refinement}}\\

\frame{\includegraphics[width=\plotwidth,]{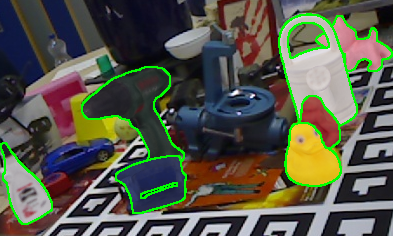}}&
\frame{\includegraphics[width=\plotwidth, ]{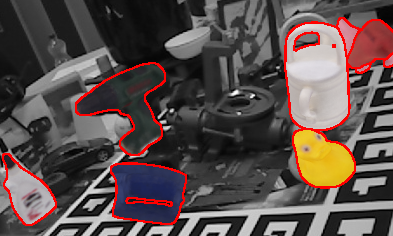}} &
\frame{\includegraphics[width=\plotwidth, ]{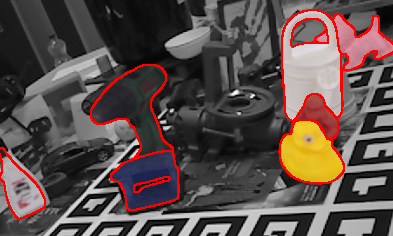}} &
\frame{\includegraphics[width=\plotwidth,]{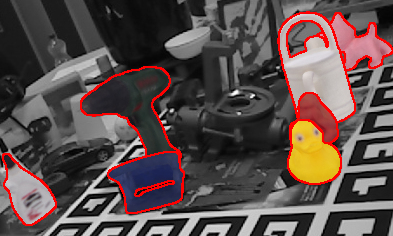}} &
\frame{\includegraphics[width=\plotwidth, ]{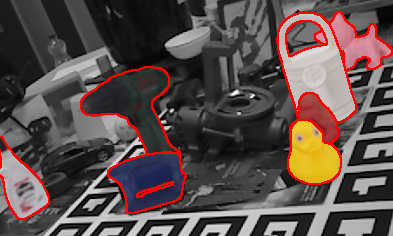}}\\ 

\frame{\includegraphics[width=\plotwidth,]{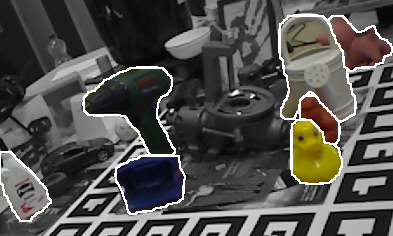}}&
\frame{\includegraphics[width=\plotwidth, ]{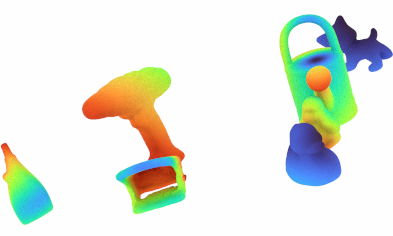}} &
\frame{\includegraphics[width=\plotwidth, ]{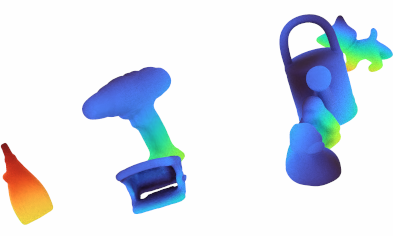}} &
\frame{\includegraphics[width=\plotwidth,]{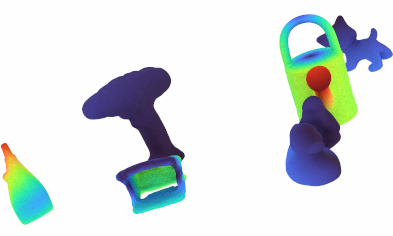}} &
\frame{\includegraphics[width=\plotwidth, ]{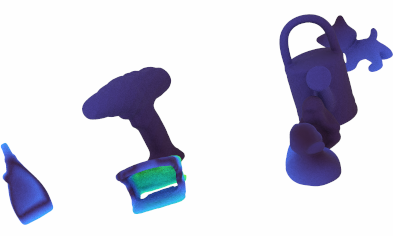}}\\

\end{tabular}
}
\vspace{-5pt}
    \captionof{figure}{ 
     \textbf{Qualitative results on LM-O~\cite{brachmann-eccv14-learning6dobjectposeestimation}}. The first column shows the ground-truth and CNOS~\cite{nguyen2023cnos} segmentation. The second and third columns show the results without refinement for both MegaPose~\cite{megapose} and our method, including depth error heatmaps at the bottom. The last two columns compare the results using the same refinement~\cite{megapose} for MegaPose~\cite{megapose} and our method. In the error heatmap, darker red indicates higher error with respect to the ground truth pose (legend: 0~cm \includegraphics[height=2mm]{images/turbo.jpg} 10~cm). As demonstrated in this figure, our method estimates a more accurate coarse pose and avoids local minima during refinement, such as with the white ``watering\_can'' object from LM-O.}
    \label{fig:qualitative}
    \vspace*{-5pt}
\end{center}
\end{figure*}

\subsection{Comparison with the State of the Art}
\label{sec:compare_SOTA}
\vspace*{-5pt}
\paragraph{Accuracy.} Table~\ref{tab:bop} compares the results of our method with those of previous work~\cite{megapose,ausserlechner2023zs6d,shugurov_osop_2022}. Across all settings, whether with or without refinement, our method consistently outperforms MegaPose while maintaining significantly faster processing times. Notably, our method significantly improves accuracy on the challenging  T-LESS, IC-BIN, and ITODD, with more than a 6\% increase in AR score for coarse pose estimation and more than a 4\% increase in AR score after refinement compared to MegaPose. 

It is important to note that although the coarse and refinement networks in MegaPose~\cite{megapose} are not trained together, they were trained to \emph{work} together: As mentioned in Section~3.2 of \cite{megapose}, the positive samples of the coarse network \textit{``are sampled from the same distribution used to generate the perturbed poses the refiner network is trained to correct''}. This pose sampling biases MegaPose's refinement process towards MegaPose's coarse estimation errors. This explains why the refinement process brings larger improvements to MegaPose than \ourmethod, in particular on TUD-L where the refinement improves MegaPose by 39.5\%  and our method only by 28.0\%. However, TUD-L represents only about 3\% of the total test data, our method still outperforms MegaPose over the 7 datasets in all settings. 
\newlength{\plotWonderwidth}
\setlength\plotWonderwidth{1.3cm}
\setlength\lineskip{0.5pt}
\setlength\tabcolsep{1.pt} 

\begin{figure}[!t]
\begin{center}
{\small
\begin{tabular}{
>{\centering\arraybackslash}m{\plotWonderwidth}
>{\centering\arraybackslash}m{\plotWonderwidth}
>{\centering\arraybackslash}m{\plotWonderwidth}
>{\centering\arraybackslash}m{\plotWonderwidth}
>{\centering\arraybackslash}m{\plotWonderwidth}
>{\centering\arraybackslash}m{\plotWonderwidth}
}
\frame{\includegraphics[width=\plotWonderwidth,]{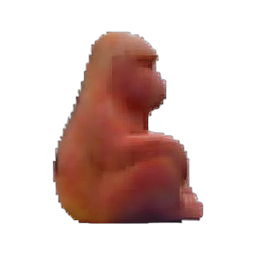}}&
\frame{\includegraphics[width=\plotWonderwidth, ]{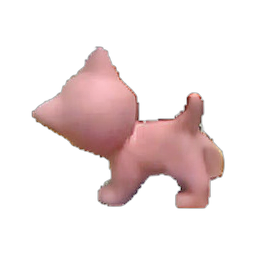}} &
\frame{\includegraphics[width=\plotWonderwidth,]{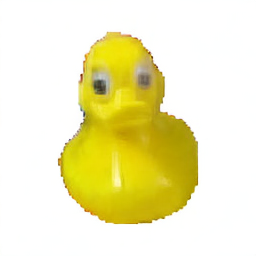}} &
\frame{\includegraphics[width=\plotWonderwidth, ]{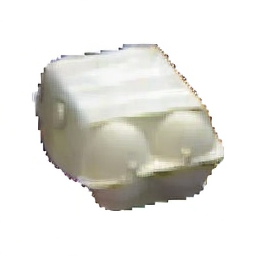}} &
\frame{\includegraphics[width=\plotWonderwidth,]{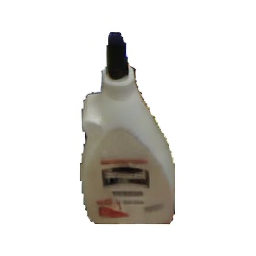}} &
\frame{\includegraphics[width=\plotWonderwidth, ]{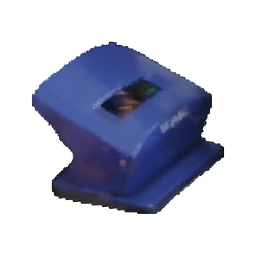}}\\ 

\frame{\includegraphics[width=\plotWonderwidth,]{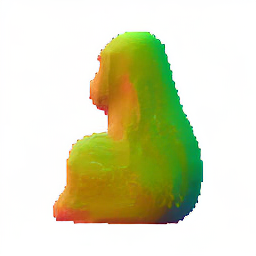}}&
\frame{\includegraphics[width=\plotWonderwidth, ]{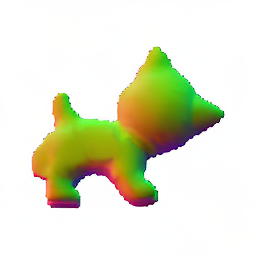}} &
\frame{\includegraphics[width=\plotWonderwidth,]{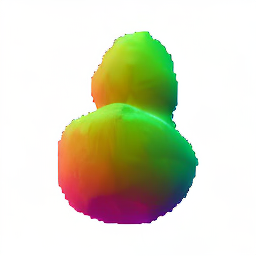}} &
\frame{\includegraphics[width=\plotWonderwidth, ]{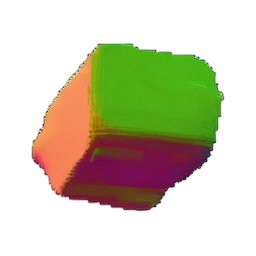}} &
\frame{\includegraphics[width=\plotWonderwidth,]{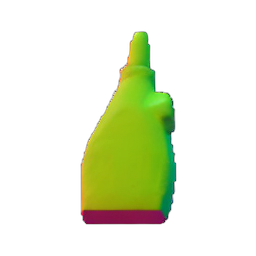}} &
\frame{\includegraphics[width=\plotWonderwidth, ]{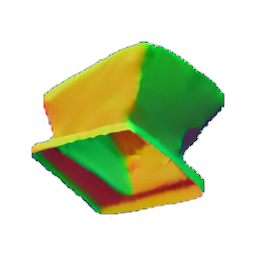}}\\ 

\end{tabular}
}

\vspace{-6pt}
    \captionof{figure}{ 
     \textbf{3D recontruction by Wonder3D~\cite{long2023wonder3d}}. The first row displays the input reference image, the second shows the predicted normal maps from the view opposite to the reference image. More visualizations are provided in the supplementary material.}
    \label{fig:wonder3d}
    \vspace*{-5pt}
\end{center}
\end{figure}
\begin{table}[t]
\centering
\setlength{\tabcolsep}{2.5pt}
\scalebox{0.83}{
\begin{tabular}{clccc|c}

\toprule  
& \multirow{2}{*}{Method}  & \multirow{2}{*}{{\makecell[b]{Detection~\cite{nguyen2023cnos}}}}  & \multicolumn{2}{c}{Single image}  & \multirow{2}{*}{{\makecell[b]{GT 3D model\\w/o refinement}}} \\ 

\cmidrule(lr){4-5}
  & & & Coarse & Refined & \\

\midrule  %
{\color{teal}\scriptsize 1}  & MegaPose~\cite{megapose} & GT 3D model & 16.3  & 25.6 & 22.9\\ %
{\color{teal}\scriptsize 2}  & \ourmethod (ours) & GT 3D model  & \bf 18.3 & \bf 27.9 & \bf 29.6\\ 
\midrule  %
{\color{teal}\scriptsize 3}  & MegaPose~\cite{megapose} & Single image & 15.4 & 25.2 & 22.7\\ {\color{teal}\scriptsize 4}  & \ourmethod (ours) & Single image & \bf 17.6 & \bf 27.2 & \bf 29.4\\ 
\bottomrule
\end{tabular}
}
\vspace{-5pt}
\caption{{\bf Results with predicted 3D models on LM-O~\cite{brachmann-eccv14-learning6dobjectposeestimation}.} We report AR score using 3D models predicted from a single reference image by Wonder3D~\cite{long2023wonder3d}. The 3D reconstruction is shown in Figure~\ref{fig:wonder3d}. Rows 3 and 4 display additional results for MegaPose and our method, where CNOS~\cite{nguyen2023cnos} is also given 3D predicted models.}

\label{tab:wonder3d}
\end{table}

Figure~\ref{fig:qualitative} shows qualitative comparisons with MegaPose~\cite{megapose} before and after refinement showcasing our more accurate pose estimates. More qualitative results are provided in the supplementary material.

\vspace*{-10pt}
\paragraph{Accuracy when using predicted 3D models.}
As shown in Table~\ref{tab:wonder3d}, our method outperforms MegaPose when using predicted 3D models. Results in Table~\ref{tab:wonder3d} implies that when no CAD model is available for an object, we can use Wonder3D to predict a 3D model from a single image, then apply GigaPose and MegaPose refinement. These results are close to GigaPose's  performance and surpass MegaPose's coarse performance when using an accurate CAD model. 

\begin{table}[t]
\centering
\setlength{\tabcolsep}{2.5pt}
\scalebox{0.85}{
\begin{tabular}{lccc}

\toprule  
\multirow{3}{*}{Method} &  \multicolumn{3}{c}{Run-time}  \\ 

\cmidrule(lr){2-4}
 &  Onboarding & Coarse pose & Refinement~\cite{megapose} \\

\midrule  %
MegaPose \cite{megapose} & 0.82 s & 1.68 s & 33 ms\\ %
\ourmethod (ours) & 11.5 s & 48 ms & 33 ms\\ 

\bottomrule
\end{tabular}
}
\vspace{-5pt}
\caption{{\bf Run-time.}  Breakdown of the average run-time for each
  stage of MegaPose~\cite{megapose} and our method on a single V100 GPU to estimate the pose \thibault{per} object (i.e., \thibault{per} detection). Our method is more than 35$\times$ faster than MegaPose for coarse pose estimation.}
\label{tab:runtime}
\end{table}

\begin{figure}[t!]
\centering
    \vspace{0.1pt}
\resizebox{0.42\textwidth}{!}{
\begin{tabular}{c}

\multicolumn{1}{c}{
\begin{tikzpicture}
\begin{customlegend}[
        legend columns=3,
        legend style={
            align=center,
            nodes={scale=1.0, transform shape},
            draw=white!80!black,
            column sep=0.1cm
        },
        legend entries={ 
            \small{T-LESS},
            \small{YCB-V},
            \small{LM-O},
        }
    ]
\addlegendimage{ultra thick, color=ForestGreen} 
\addlegendimage{ultra thick, color=purple}
\addlegendimage{ultra thick, color=YellowOrange}
\end{customlegend}
\end{tikzpicture}}
\\

\multicolumn{1}{c}{
\begin{tikzpicture}
\begin{customlegend}[
        legend columns=2,
        legend style={
            align=center,
            nodes={scale=1.0, transform shape},
            draw=white!80!black,
            column sep=0.1cm
        },
        legend entries={
            \small{MegaPose},
            \small{Ours}
        }
    ]
\addlegendimage{ultra thick, color=gray, dashed}
\addlegendimage{ultra thick, color=gray, mark=+}
\end{customlegend}
\end{tikzpicture}}
\\
\begin{tikzpicture}
  \tikzstyle{every node}=[font=\small]
    \begin{axis}[
        xmin=0.1,
        xmax=1,
        ymin=0,
        ymax=35,
        width=8.0cm,
        height=6.0cm,
        font=\footnotesize,
        xtick={0.1, 0.3, 0.5, 0.7, 0.9},
        xlabel=IoU,
        ytick={10, 20, 30},
        ylabel=AP ($\%$),
        ylabel style={yshift=-5pt},
        inner sep=3pt,
        label style={font=\small},
        tick label style={font=\small},
        legend pos=south east,
        grid=major,
        legend style={nodes={scale=0.8, transform shape}},
        legend cell align={center}
    ]
    \addplot[thick, color=ForestGreen, dashed]  table [y=y,x=x]{tikz/data/tless_megapose.txt};
    \addplot[thick, color=ForestGreen, mark=+]  table [y=y,x=x]{tikz/data/tless_ours.txt};
    \legend{};

    \addplot[thick, color=purple, dashed]  table [y=y,x=x]{tikz/data/ycbv_megapose.txt};
    \addplot[thick, color=purple, mark=+]  table [y=y,x=x]{tikz/data/ycbv_ours.txt};
    
    \addplot[thick, color=YellowOrange, dashed]  table [y=y,x=x]{tikz/data/lmo_megapose.txt};
    \addplot[thick, color=YellowOrange, mark=+]  table [y=y,x=x]{tikz/data/lmo_ours.txt};
    
    \end{axis}
\end{tikzpicture}
\end{tabular}
}
    \vspace{-8pt}
    \caption{\textbf{Robustness to segmentation errors.} We analyze the performance of MegaPose and our method under various levels of segmentation errors, defined by the IoU between the predicted masks from CNOS~\cite{nguyen2023cnos} and the ground-truth masks. Our method demonstrates much higher stability in AP across all IoU thresholds than MegaPose, showing its robustness against segmentation errors. The improvement is more limited on LM-O because of the small appearance size of the objects especially after occlusions.}    
    \label{fig:occlusion}
\end{figure}
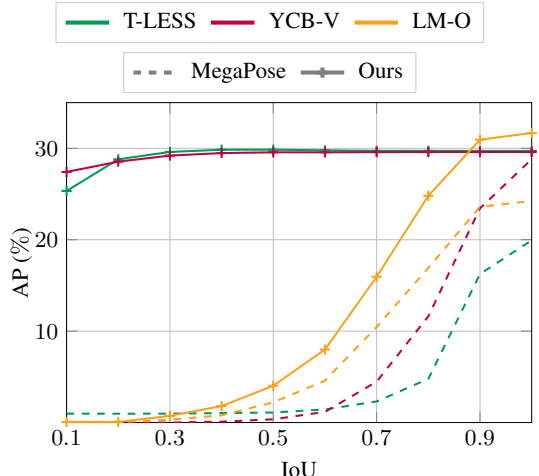

\vspace*{-12pt}
\paragraph{Run-time.} We report the speed of \ourmethod in Table~\ref{tab:bop}~(rightmost column) following the BOP evaluation protocol. It measures the total processing time \textit{per image} averaged over the datasets including the time taken by CNOS~\cite{nguyen2023cnos} to segment each object, the time to estimate the object pose for all detections, and the refinement time  if applicable. 

Table~\ref{tab:runtime} gives a  breakdown of the run-time \textit{per detection} for each stage of MegaPose and of our method. Our method takes only 48 ms for coarse pose estimation, more than 38x faster than the 1.68 seconds taken by MegaPose. This improvement can be attributed to our sublinear nearest neighbor search, significantly faster than feed-forwarding each of the 576 input-template pairs as done in MegaPose.


\vspace{-15pt}
\paragraph{Robustness to segmentation errors.} To demonstrate the robustness of our method, we analyze  its performance under various levels of segmentation errors on three standard datasets: LM-O~\cite{brachmann-eccv14-learning6dobjectposeestimation}, T-LESS~\cite{hodan-wacv17-tless}, and YCB-V~\cite{Xiang2018-dv}. We use the ground-truth masks to classify the segmentation errors produced by CNOS's segmentation~\cite{nguyen2023cnos} using the Intersection over Union (IoU) metric. For each IoU threshold, we retain only the input masks from CNOS that matched the ground-truth masks with an IoU smaller than this threshold and evaluate the AR score for coarse pose estimation.

As shown in Figure~\ref{fig:occlusion}, our method has a stable AR score across all IoU thresholds for both T-LESS and YCB-V, in contrast with MegaPose, which yields high scores primarily for high IoU thresholds only. 

\begin{figure}[!t]
\setlength\plotWonderwidth{1.6cm}

\begin{center}
{\small
\begin{tabular}{ccc}
\includegraphics[height=\plotWonderwidth,]{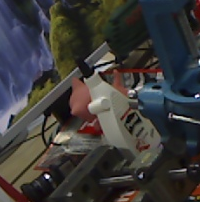}&
\includegraphics[height=\plotWonderwidth, ]{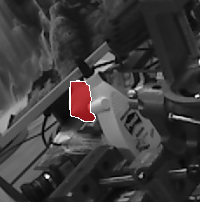} &
\frame{\includegraphics[height=\plotWonderwidth,]{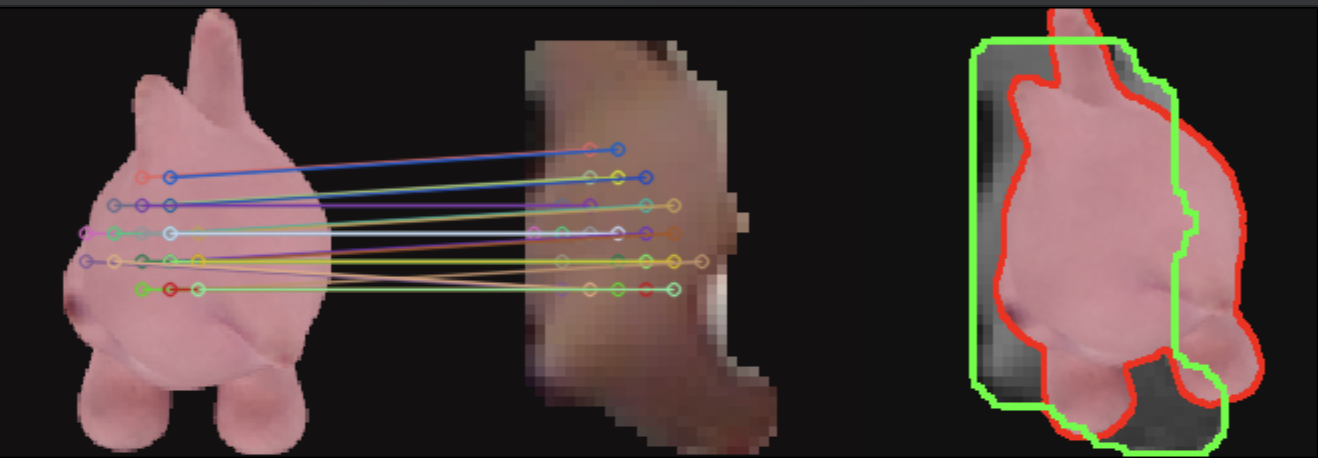}}\\ 
\scriptsize{Input RGB} & \scriptsize{Segmentation~\cite{nguyen2023cnos}} & \scriptsize{2D-to-2D correspondences \;\;\;\; Prediction}
\end{tabular}
}
\captionof{figure}{ 
\thibault{\textbf{Failure case.} The ``Cat'' object of LM-O~\cite{brachmann-eccv14-learning6dobjectposeestimation} is not retrieved correctly here because of the small size of its segment and the low-fidelity CAD models, resulting in outlier matches.}}
 \label{fig:failureLMO}
 \vspace*{-5pt}

\end{center}
\end{figure}
\vspace*{-5pt}
\subsection{Failure cases}
\vspace*{-5pt}
\nguyen{Figure~\ref{fig:occlusion} also shows that the AR score stability is less important in the case of LM-O where many challenging conditions are present, including heavy occlusions, low resolution segmentation, and low-fidelity CAD models. We show in Figure~\ref{fig:failureLMO} this failure case on the ``Cat'' object of LM-O.}

\vspace*{-5pt}
\subsection{Ablation Study}
\vspace*{-5pt}
In Table~\ref{tab:ablation}, we present several ablation evaluations on the three standard datasets LM-O~\cite{brachmann-eccv14-learning6dobjectposeestimation}, T-LESS~\cite{hodan-wacv17-tless}, and YCB-V~\cite{Xiang2018-dv}. Our results are in \colorbox{gray!20}{Row 5}. 
\label{sec:ablation}
\begin{table}[t]
\centering
\setlength{\tabcolsep}{2.5pt}
\scalebox{0.85}{
\begin{tabular}{lcccc cccc}
\toprule  
& \multirow{3}{*}{{\makecell[b]{Fine tune \\ $\invariantNet$}}} & \multirow{3}{*}{{\makecell[b]{Templates\\in-plane}}} & \multicolumn{2}{c}{P$n$P} & \multirow{3}{*}{\textsc{lm-o}} & \multirow{3}{*}{\textsc{t-less}} & \multirow{3}{*}{\textsc{ycb-v}} & \multirow{3}{*}{\textsc{Mean}} \\ 
\cmidrule(lr){4-5}
& & & Type & $n$ \\ 

\midrule  %

{\color{teal}\scriptsize 1} & $\redxmark$  & $\redxmark$ & 2D-to-2D & 1  & 20.1 & 19.3 &  17.7 & 19.0 \\

\midrule  %
{\color{teal}\scriptsize 2} & $\greencheckmark$ & $\greencheckmark$ & 2D-to-2D & 1 & 23.3 & 21.1 &  22.1 & 22.1 \\
\midrule  %
{\color{teal}\scriptsize 3} & $\greencheckmark$ &  $\redxmark$ & 3D-to-2D & 4  & 28.0 & 25.3 &  26.3 & 26.5 \\ 

{\color{teal}\scriptsize 4} & $\greencheckmark$ &  $\redxmark$ & 2D-to-2D & 2 &  \bf 30.0 & 25.6 & 26.0 & 27.2 \\

\rowcolor{gray!20}{\color{teal}\scriptsize 5} & $\greencheckmark$ &  $\redxmark$ & 2D-to-2D & 1 & 29.6 & \bf 26.4 &  \bf 27.8 & \bf 27.9\\ %

\midrule  
\rowcolor{yellow!20}{\color{teal}\scriptsize 6} & $\greencheckmark$ &  $\redxmark$ & 2D-to-2D & 1 & \textit{30.1} & \textit{27.1} &  \textit{28.4} & \textit{28.5}\\ %




\bottomrule
\end{tabular}
}
\caption{{\bf Ablation study.} 
We report the AR score of different settings of our method including:
without fine-tuning $\invariantNet$ in Row 1, estimating in-plane rotation with dense 3DoF templates in Row 2, different ``P$n$P'' variants in Rows 3 and 4. The results of the complete method are on \colorbox{gray!20}{Row 5}. \nguyen{We show in \colorbox{yellow!20}{Row 6} our results using the same 576 templates as in MegaPose~\cite{megapose}.} See Section~\ref{sec:ablation}.}
\vspace*{-2pt}


\label{tab:ablation}
\end{table}

\vspace*{-5pt}
\paragraph{Fine-tuning $\invariantNet$.} Row 1 of Table~\ref{tab:ablation} presents the results of using the DINOv2 features~\cite{oquab2023dinov2} without fine-tuning $\invariantNet$. As shown in Row 5, fine-tuning significantly improves template-correspondences, leading to a 8.9\% increase in AR score.

\vspace*{-3pt}
\paragraph{Estimating in-plane rotation with templates.} Row 2 of Table~\ref{tab:ablation} shows in-plane rotation estimation results using templates by dividing in-plane angle into 36 bins of 10 degrees, yielding 5832 templates per object. This approach decreases the AR score by 5.8\% compared to direct predictions with $\variantNet$ and $\variantMLP$ in Row 5, underscoring the effectiveness of our hybrid template-patch correspondence approach.

\vspace*{-3pt}
\paragraph{2D-to-2D vs 3D-to-2D correspondences.}
In Row 3, we introduce a ``3D-to-2D correspondence'' variant by replacing the 2D locations of the matched patches in the template with their 3D counterparts obtained from the template depth map. We then estimate the complete 6D object pose using the eP$n$P algorithm~\cite{lepetit2009ep}  implemented in OpenCV~\cite{opencv_library}. Furthermore, in Row 4, we present a two-``2D-to-2D correspondences''
variant, where the scale and in-plane rotation are computed using a 2D variant of the Kabsch algorithm (more details are given in the supplementary material). Our single-correspondence approach in Row 5 is more effective at exploiting patch correspondences for estimating scale and in-plane rotation directly.

\vspace*{-3pt}
\nguyen{\paragraph{Number of templates.} In Row 6, we present our results using the same 576 templates as MegaPose~\cite{megapose}. This  improves by only 0.6\% the AR score compared to using 162 templates (Row 5). This confirms that the correspondences also allows to decrease the memory footprint of the templates without hurting the accuracy.}
\vspace*{-3pt}

\vspace*{4pt}
\section{Conclusion}

We presented  \ourmethod, an efficient method for the 6D coarse pose estimation of novel objects. \nguyen{It stands out for its significant speed, robustness, and accuracy compared to existing methods, and can be seamlessly integrated with any refinement methods. We hope that \ourmethod will make real-time accurate pose estimation of novel objects practical.} \thibault{We discuss avenues for future work in the supplementary}.

\newpage
\textbf{\Large{Supplementary Material}}
\section{Ground-truth 2D-to-2D correspondences}
\label{sec:generate_correspondence}

As discussed in Section 3.2 of the main paper, we use the ground-truth 3D information of training sets provided by the BOP challenge~\cite{sundermeyer2023bop}, originally sourced from MegaPose~\cite{megapose} to create the 2D-to-2D correspondences for training.

For each 2D location $i$---the 2D center for a patch of size 14$\times$14 of the query image---we aim to identify its corresponding location $i^*$ in the nearest template. We achieve this through a straightforward re-projection process. For each 2D center in the query image, we first calculate its 3D counterpart using the query depth map and camera intrinsics. We then transform this 3D point into the camera view of the nearest template using the ground-truth relative pose, and re-project this 3D point into the template using template camera intrinsics. If the re-projected 2D location falls inside the template mask, we identify the nearest patch $i^*$ among all patches within the template mask, as the corresponding location for the input query patch $i$. 

We reverse the roles of the query and the template, then use the same process to establish 2D-to-2D correspondences for each patch of the template.

\begin{figure}[!t]
    \begin{center}
    \includegraphics[width=\linewidth]{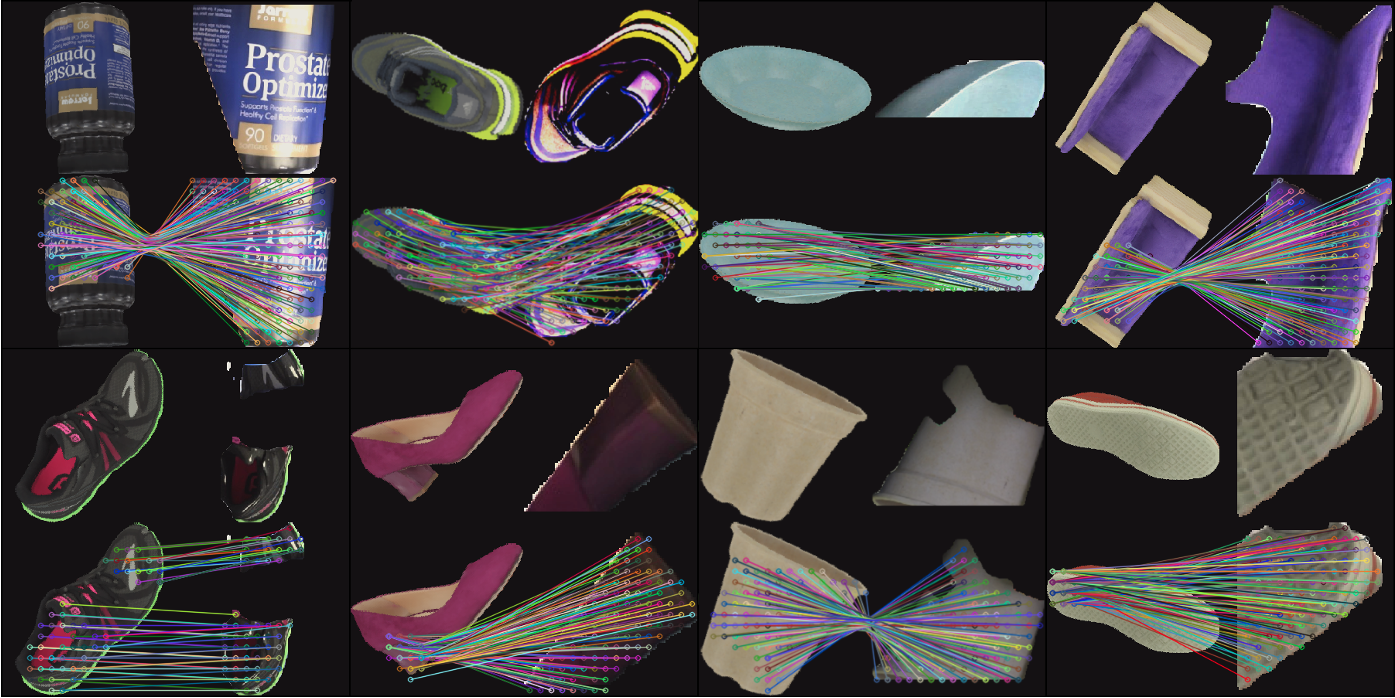}
    \end{center}
    \vspace{-12pt}
    \caption{
        \label{fig:training_sample}
        {\bf Training samples.} For each training sample (2$\times$2 images), we show the template image $\template_k$ (top left), the query image $\query_k$ (top right) and the 2D-to-2D correspondences (bottom) as discussed in Section~\ref{sec:generate_correspondence}. }
\end{figure}

We use color augmentation to close the real-synthetic domain gap as done in~\cite{megapose}, including: Gaussian blur, contrast, brightness, colors and sharpness filters from the Pillow library~\cite{pillow}. We show in Figure~\ref{fig:training_sample} eight training samples created from this process.


\section{Recovering a 6D object pose}
\label{sec:recovering_6d}

For simplicity, in this section, we denote the input template and the input testing image \textit{before} the processing step (i.e., scaling, cropping, and padding) as $\template$ and $\query$, respectively.

As mentioned in Section 3 of the main paper, we decompose the 6D object pose into out-of-plane rotation $\Routplane$ and the affine transform $\affineTransform$ (including in-plane rotation $\alpha$, 2D scale and 2D translation), which transforms the processed template to the processed query. We detail below how we can recover the 6D object pose from $\Routplane$ and $\affineTransform$.

First, the 3~DoF for the rotation can be recovered by simply combining the out-of-plane rotation $\Routplane$, annotated alongside the nearest template, with the in-plane rotation, $\alpha$, as predicted by the network $\variantNet$:
\begin{equation}
\begin{aligned}
\bR &= \bR_{\alpha}  \Routplane \\
    &= \begin{bmatrix}
    \cos(\alpha) & -\sin(\alpha) & 0 \\
    \sin(\alpha) & \cos(\alpha) & 0 \\
    0 & 0 & 1 \\
\end{bmatrix} \>  \Routplane  \> .
\end{aligned}
\label{eq:rotation}
\end{equation}
Recovering 3~DoF for object translation in the test image involves additional transformations, $\croppingQuery$ and $\croppingTemplate$, for scaling, cropping, and padding the input testing image $\query$ and the input template $\template$, respectively. These transformations are used to standardize the input images to a fixed size of 224$\times$224, and can be defined as:
\begin{equation}
\begin{aligned}
\croppingTemplate &= \begin{bmatrix}
    \mathbf{s} & 0 & \mathbf{t}_{x} \\
    0 & \mathbf{s} & \mathbf{t}_{y} \\
    0 & 0 & 1 \\
\end{bmatrix} \> ,
\end{aligned}
\label{eq:cropping}
\end{equation}
where $\mathbf{s}$ is the scaling factor, and $\left[\mathbf{t}_{x}, \mathbf{t}_{y}\right]$ is the 2D translation, created from the scaling, cropping, and padding applied to the template. Similarly, we can define $\croppingQuery$, the transformation applied to the input testing image $\query$.

\begin{figure}[!t]
    \begin{center}
    \frame{\includegraphics[width=0.95\linewidth]{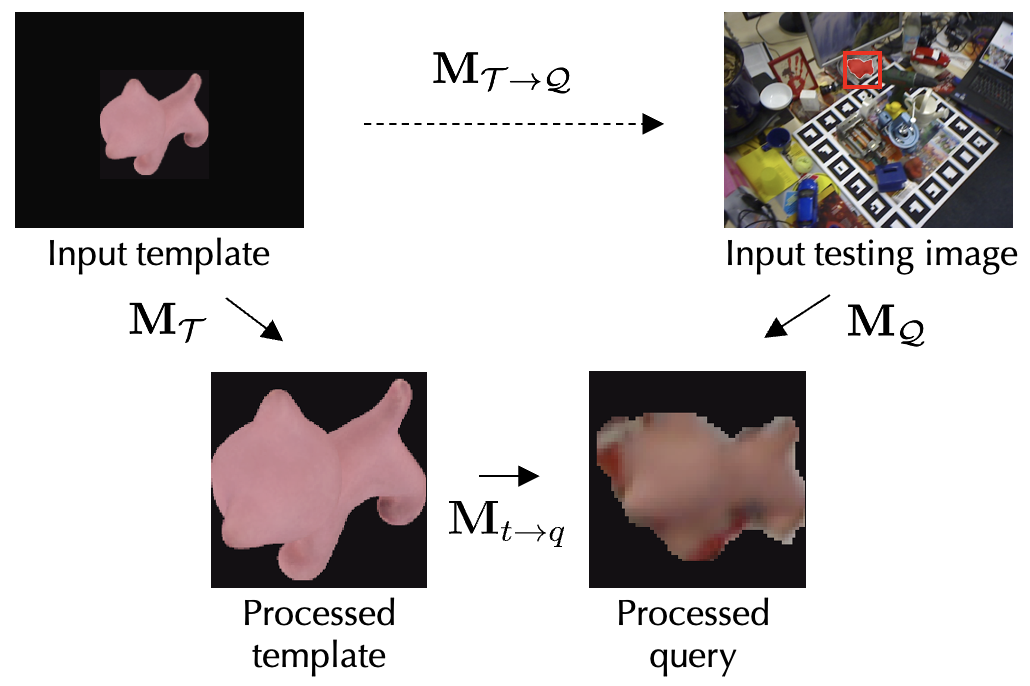}}
    \end{center}
    \vspace{-12pt}
    \caption{
        \label{fig:all_trasnformations}
        {\bf Transformations from template $\template$ to query $\query$.} We show all the transformations has been applied to transform the template $\template$ to the input testing image $\query$, as discussed in Section~\ref{sec:recovering_6d}. }
\end{figure}

As shown in Figure~\ref{fig:all_trasnformations}, the transformation $\fullAffineTransform$ transforming a 2D point in the template $\template$ to its 2D corresponding point in the query $\query$ can be defined explicitly as:
\begin{equation}
\begin{aligned}
\fullAffineTransform &= \croppingTemplate  \affineTransform  \croppingQuery^{-1} \> .
\end{aligned}
\label{eq:template2query}
\end{equation}

We can therefore use the transformation $\fullAffineTransform$ to recover the 2D projection of the object's translation in the query image, $\left[\mathbf{c}_{\mathcal{Q}, x}, \mathbf{c}_{\mathcal{Q}, y}\right]$ given the 2D projection of object's translation in the template, $\left[\mathbf{c}_{\mathcal{T}, x}, \mathbf{c}_{\mathcal{T}, y}\right]$:
\begin{equation}
\begin{aligned}
\begin{bmatrix}
    \mathbf{c}_{\mathcal{Q}, x} \\ \mathbf{c}_{\mathcal{Q}, y} \\
    1 \\
    \end{bmatrix} 
&= \fullAffineTransform  \begin{bmatrix}
    \mathbf{c}_{\mathcal{T}, x} \\ \mathbf{c}_{\mathcal{T}, y} \\
    1 \\
    \end{bmatrix} \> .
\end{aligned}
\label{eq:projectionQuery}
\end{equation}
The only missing degree is the object's translation of the query image in Z axis, $\mathbf{t}_{\query, z}$, which can be deduced from $\mathbf{t}_{\template, z}$, the object's translation of the template in Z axis, $\fullAffineTransform$ and the focal ratio using the following formula:
\begin{equation}
\begin{aligned}
    \mathbf{t}_{\query, z}&= \mathbf{t}_{\template, z} \times \frac{1}{\text{scale}\left(\fullAffineTransform\right)} \times \frac{\mathbf{f}_{\query}}{\mathbf{f}_{\template}} \> ,
\end{aligned}
\label{eq:translation_z}
\end{equation}
where $\text{scale}\left(\fullAffineTransform\right)$ is the 2D scale in $\fullAffineTransform$, which is equal to the norm of first column of $\fullAffineTransform$, and $\mathbf{f}_{(.)}$ is the focal length.

Finally, we calculate the object's translation in the query image $\left[\mathbf{t}_{\mathcal{Q}, x}, \mathbf{t}_{\mathcal{Q}, y}, \mathbf{t}_{\mathcal{Q}, z}\right]$ using the query camera intrinsic $\mathbf{K}_{\mathcal{Q}}$:
\begin{equation}
\begin{aligned}
\begin{bmatrix}
    \mathbf{t}_{\mathcal{Q}, x} \\ \mathbf{t}_{\mathcal{Q}, y} \\
    \mathbf{t}_{\mathcal{Q}, z} \\
    \end{bmatrix} 
&= \mathbf{t}_{\mathcal{Q}, z} \times \left( \mathbf{K}_{\mathcal{Q}}^{-1} \begin{bmatrix}
    \mathbf{c}_{\mathcal{Q}, x} \\ \mathbf{c}_{\mathcal{Q}, y} \\
    1 \\
    \end{bmatrix} \right) \> .
\end{aligned}
\label{eq:projectionQuery}
\end{equation}

\section{``2D'' version of the Kabsch algorithm}
\label{sec:kabsch2d}

The classic Kabsch algorithm~\cite{Kabsch1976ASF} has been commonly used for points in a three-dimensional space. In this work, we use a ``2D'' version for two-dimensional space. This allows us to recover the affine transformation $\affineTransform$, including the 2D scale $\mathbf{s}$, in-plane rotation $\bR_{\alpha}$, and 2D translation $\mathbf{t}$ from two 2D-to-2D correspondences.

Let denote $\{({\mathbf{p}_{\template}^1, \mathbf{p}_{\query}^1}), ({\mathbf{p}_{\template}^2, \mathbf{p}_{\query}^2}) \}$ two correspondences that we obtain from the nearest neighbor search with the features of $\invariantNet$. Our goal is to find $\{ \mathbf{s}, \bR_{\alpha}, \mathbf{t}\}$ which transforms $\mathbf{p}_{\template}^1$ to $\mathbf{p}_{\query}^1$, and $\mathbf{p}_{\template}^2$ to $\mathbf{p}_{\query}^2$:
\begin{equation}
\begin{aligned}
\mathbf{p}_{\query}^1 = \mathbf{s} \times \bR_{\alpha} \mathbf{p}_{\template}^1 + \mathbf{t} \> ,\\
\mathbf{p}_{\query}^2 = \mathbf{s} \times \bR_{\alpha} \mathbf{p}_{\template}^2 + \mathbf{t} \> .\\ 
\end{aligned}
\label{eq:goalKabsch2d}
\end{equation}

First, we calculate the scale $\mathbf{s}$ from the size of two vectors $\mathbf{p}_{\query}^2 -\mathbf{p}_{\query}^1$ and $\mathbf{p}_{\template}^2-\mathbf{p}_{\template}^1$:
\begin{equation}
\begin{aligned}
\mathbf{s} = \frac{||\mathbf{p}_{\query}^2 -\mathbf{p}_{\query}^1 ||}{||\mathbf{p}_{\template}^2-\mathbf{p}_{\template}^1 ||} \> .
\end{aligned}
\label{eq:KabschScale}
\end{equation}

The rotation matrix $\bR_{\alpha}$ is composed of $\cos(\alpha)$ and $\sin(\alpha)$, and is defined as:
\begin{equation}
\begin{aligned}
\bR_{\alpha} = \begin{bmatrix}
    \cos(\alpha) & -\sin(\alpha)\\
    \sin(\alpha) & \cos(\alpha) \\
\end{bmatrix} \> ,
\end{aligned}
\label{eq:KabschRotation}
\end{equation}
where $\cos(\alpha)$ and $\sin(\alpha)$ are the dot product and cross product respectively of vectors $\mathbf{p}_{\query}^2 -\mathbf{p}_{\query}^1$ and $\mathbf{p}_{\template}^2-\mathbf{p}_{\template}^1$:
\begin{equation}
\begin{aligned}
\cos(\alpha) = \frac{\left(\mathbf{p}_{\template}^2 -\mathbf{p}_{\template}^1\right)^{T} . \left(\mathbf{p}_{\query}^2 -\mathbf{p}_{\query}^1\right)}{||\mathbf{p}_{\template}^2-\mathbf{p}_{\template}^1 ||. ||\mathbf{p}_{\query}^2 -\mathbf{p}_{\query}^1 ||} \> ,
\end{aligned}
\label{eq:cosAlpha}
\end{equation}
\begin{equation}
\begin{aligned}
\sin(\alpha) = \frac{\left(\mathbf{p}_{\template}^2 -\mathbf{p}_{\template}^1\right)^{T} \wedge \left(\mathbf{p}_{\query}^2 -\mathbf{p}_{\query}^1\right)}{||\mathbf{p}_{\template}^2-\mathbf{p}_{\template}^1 ||. ||\mathbf{p}_{\query}^2 -\mathbf{p}_{\query}^1 ||} \> .
\end{aligned}
\label{eq:sinAlpha}
\end{equation}

Given the predicted scale $\mathbf{s}$ and rotation matrix $\bR_{\alpha}$, we can deduce translation $\mathbf{t}$:
\begin{equation}
\begin{aligned}
\mathbf{t} =  \frac{1}{2} \left[\left( \mathbf{p}_{\query}^1 - \mathbf{s} \times \bR_{\alpha} \mathbf{p}_{\template}^1 \right) + \left( \mathbf{p}_{\query}^2 - \mathbf{s} \times \bR_{\alpha} \mathbf{p}_{\template}^2 \right) \right] \> .
\end{aligned}
\label{eq:KabschTranslation}
\end{equation}

\section{Additional results}
\label{sec:addition_results}

\setlength\plotWonderwidth{1.33cm}
\setlength\lineskip{1.5pt}
\setlength\tabcolsep{1.pt} 
\definecolor{darkgreen}{rgb}{0, 0.5, 0}

\begin{figure}[!t]
\begin{center}
{\small
\begin{tabular}{
>{\centering\arraybackslash}m{\plotWonderwidth}
>{\centering\arraybackslash}m{\plotWonderwidth}
>{\centering\arraybackslash}m{\plotWonderwidth}
>{\centering\arraybackslash}m{\plotWonderwidth}
>{\centering\arraybackslash}m{\plotWonderwidth}
>{\centering\arraybackslash}m{\plotWonderwidth}
}
{\color{green}\fbox{\includegraphics[width=0.8\plotWonderwidth,]{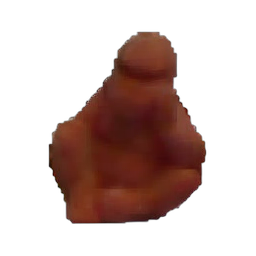}}}&
{\color{black}\fbox{\includegraphics[width=0.8\plotWonderwidth, ]{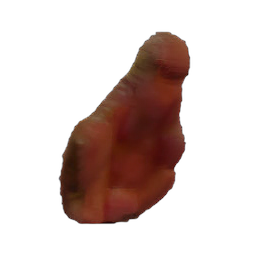}}} &
{\color{black}\fbox{\includegraphics[width=0.8\plotWonderwidth,]{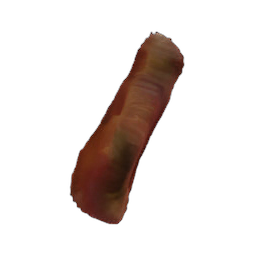}}} &
{\color{black}\fbox{\includegraphics[width=0.8\plotWonderwidth, ]{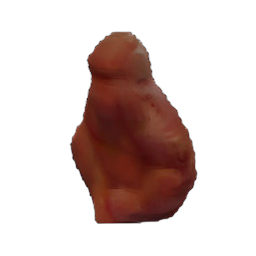}}} &
{\color{black}\fbox{\includegraphics[width=0.8\plotWonderwidth,]{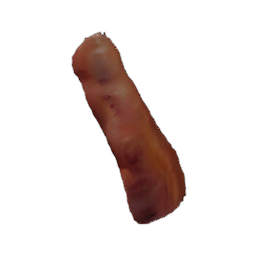}}} &
{\color{black}\fbox{\includegraphics[width=0.8\plotWonderwidth, ]{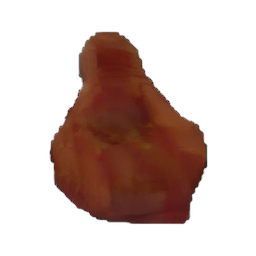}}}\\

\noalign{\vspace{0.5ex}}
{\color{black}\fbox{\includegraphics[width=0.8\plotWonderwidth,]{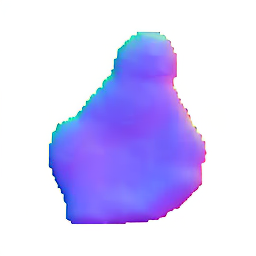}}}&
{\color{black}\fbox{\includegraphics[width=0.8\plotWonderwidth, ]{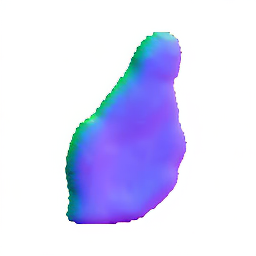}}} &
{\color{black}\fbox{\includegraphics[width=0.8\plotWonderwidth,]{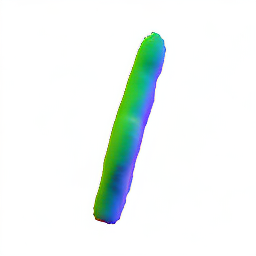}}} &
{\color{black}\fbox{\includegraphics[width=0.8\plotWonderwidth, ]{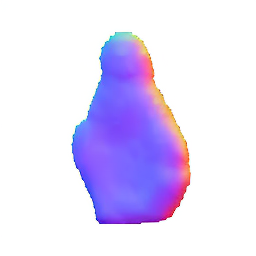}}} &
{\color{black}\fbox{\includegraphics[width=0.8\plotWonderwidth,]{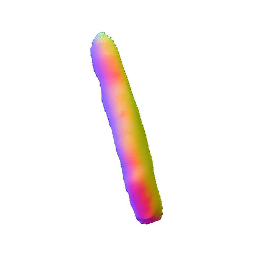}}} &
{\color{black}\fbox{\includegraphics[width=0.8\plotWonderwidth, ]{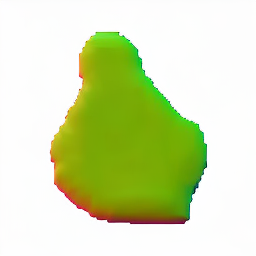}}}\\

\noalign{\vspace{0.5ex}}
{\color{green}\fbox{\includegraphics[width=0.8\plotWonderwidth,]{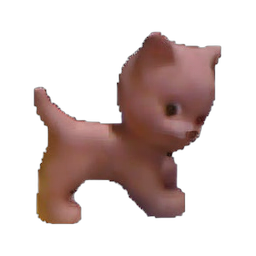}}}&
{\color{black}\fbox{\includegraphics[width=0.8\plotWonderwidth, ]{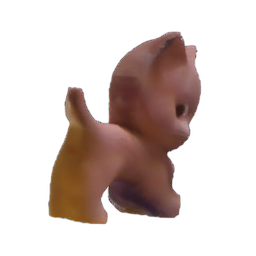}}} &
{\color{black}\fbox{\includegraphics[width=0.8\plotWonderwidth,]{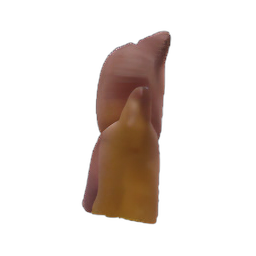}}} &
{\color{black}\fbox{\includegraphics[width=0.8\plotWonderwidth, ]{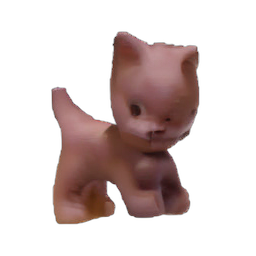}}} &
{\color{black}\fbox{\includegraphics[width=0.8\plotWonderwidth,]{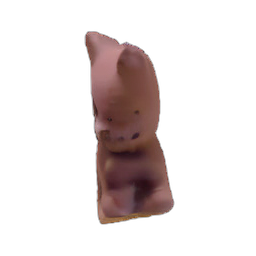}}} &
{\color{black}\fbox{\includegraphics[width=0.8\plotWonderwidth, ]{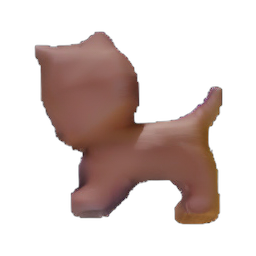}}}\\ 

\noalign{\vspace{0.5ex}}
{\color{black}\fbox{\includegraphics[width=0.8\plotWonderwidth,]{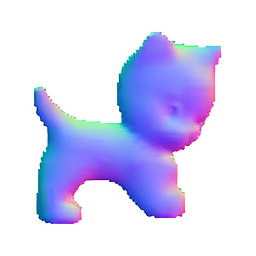}}}&
{\color{black}\fbox{\includegraphics[width=0.8\plotWonderwidth, ]{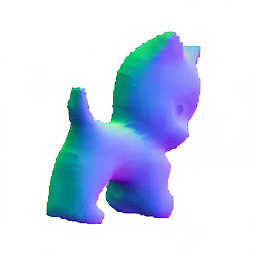}}} &
{\color{black}\fbox{\includegraphics[width=0.8\plotWonderwidth,]{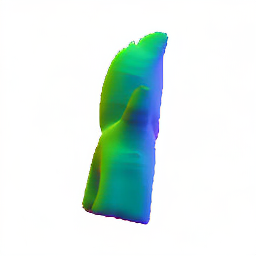}}} &
{\color{black}\fbox{\includegraphics[width=0.8\plotWonderwidth, ]{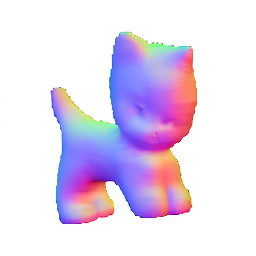}}} &
{\color{black}\fbox{\includegraphics[width=0.8\plotWonderwidth,]{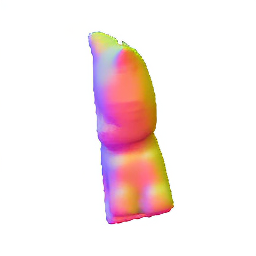}}} &
{\color{black}\fbox{\includegraphics[width=0.8\plotWonderwidth, ]{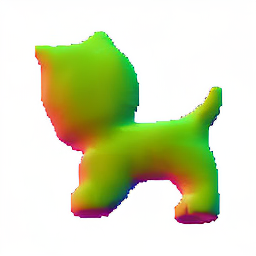}}}\\ 
\scriptsize{Front} & \scriptsize{Front left} & \scriptsize{Left} & \scriptsize{Front-right} & \scriptsize{Right} & \scriptsize{Back}
\end{tabular}
}

    \captionof{figure}{ 
     \textbf{Failure cases of Wonder3D~\cite{long2023wonder3d}}.
     We present common failure cases of the Wonder3D with the ``ape'' and ``cat'' objects from the LM-O dataset~\cite{brachmann-eccv14-learning6dobjectposeestimation}. For each sample (2$\times$6 images), we show the input image outlined in {\color{darkgreen}{green}} (top left), the predicted rgb (second to last column of the first row), and the second row shows the corresponding predicted normals. These objects appear ``flat'' when viewed from novel angles. }
    \label{fig:failureCases}
    \vspace*{-5pt}
\end{center}
\end{figure}
\subsection{Using 3D models predicted by Wonder3D}
\label{sec:reconstruction_wonder3d}
\newlength{\plotWonderwidthSuppMat}
\setlength\plotWonderwidthSuppMat{1.33cm}
\setlength\lineskip{1.5pt}
\setlength\tabcolsep{1.pt} 
\definecolor{darkgreen}{rgb}{0, 0.5, 0}

\begin{figure*}[!t]
\begin{center}
{\small
\begin{tabular}{
>{\centering\arraybackslash}m{\plotWonderwidthSuppMat}
>{\centering\arraybackslash}m{\plotWonderwidthSuppMat}
>{\centering\arraybackslash}m{\plotWonderwidthSuppMat}
>{\centering\arraybackslash}m{\plotWonderwidthSuppMat}
>{\centering\arraybackslash}m{\plotWonderwidthSuppMat}
>{\centering\arraybackslash}m{\plotWonderwidthSuppMat}
>{\centering\arraybackslash}m{0.5cm}
>{\centering\arraybackslash}m{\plotWonderwidthSuppMat}
>{\centering\arraybackslash}m{\plotWonderwidthSuppMat}
>{\centering\arraybackslash}m{\plotWonderwidthSuppMat}
>{\centering\arraybackslash}m{\plotWonderwidthSuppMat}
>{\centering\arraybackslash}m{\plotWonderwidthSuppMat}
>{\centering\arraybackslash}m{\plotWonderwidthSuppMat}
}
{\color{green}\fbox{\includegraphics[width=0.8\plotWonderwidthSuppMat,]{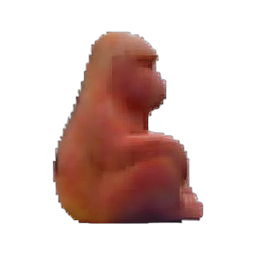}}}&
{\color{black}\fbox{\includegraphics[width=0.8\plotWonderwidthSuppMat]{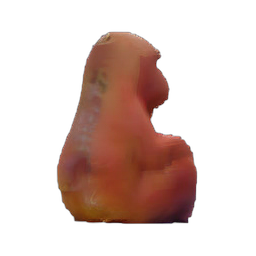}}} &
{\color{black}\fbox{\includegraphics[width=0.8\plotWonderwidthSuppMat,]{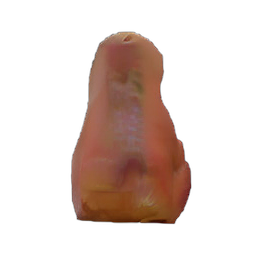}}} &
{\color{black}\fbox{\includegraphics[width=0.8\plotWonderwidthSuppMat]{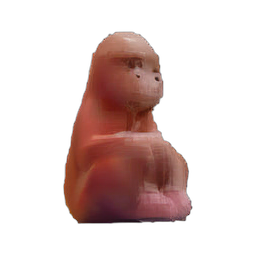}}} &
{\color{black}\fbox{\includegraphics[width=0.8\plotWonderwidthSuppMat,]{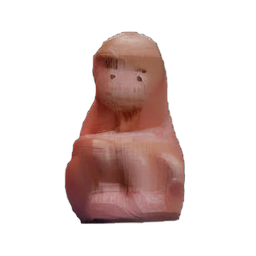}}} &
{\color{black}\fbox{\includegraphics[width=0.8\plotWonderwidthSuppMat]{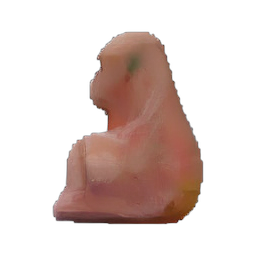}}} & &

{\color{green}\fbox{\includegraphics[width=0.8\plotWonderwidthSuppMat,]{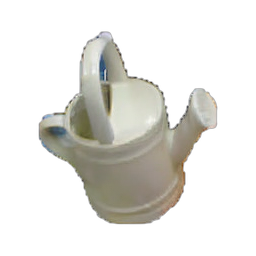}}}&
{\color{black}\fbox{\includegraphics[width=0.8\plotWonderwidthSuppMat]{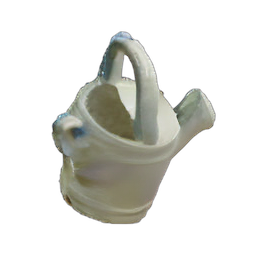}}} &
{\color{black}\fbox{\includegraphics[width=0.8\plotWonderwidthSuppMat,]{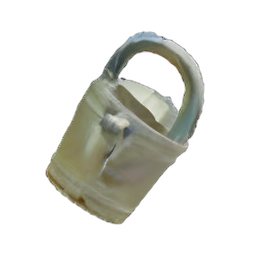}}} &
{\color{black}\fbox{\includegraphics[width=0.8\plotWonderwidthSuppMat]{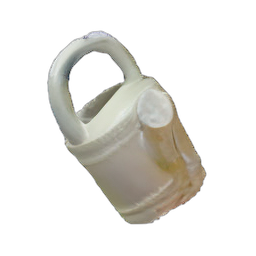}}} &
{\color{black}\fbox{\includegraphics[width=0.8\plotWonderwidthSuppMat,]{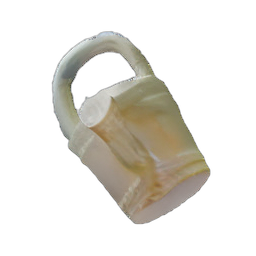}}} &
{\color{black}\fbox{\includegraphics[width=0.8\plotWonderwidthSuppMat]{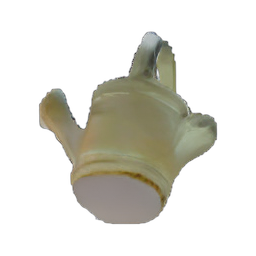}}} \\

\noalign{\vspace{0.5ex}}
{\color{black}\fbox{\includegraphics[width=0.8\plotWonderwidthSuppMat,]{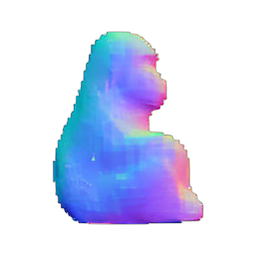}}}&
{\color{black}\fbox{\includegraphics[width=0.8\plotWonderwidthSuppMat]{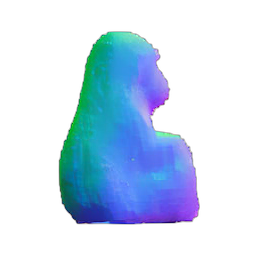}}} &
{\color{black}\fbox{\includegraphics[width=0.8\plotWonderwidthSuppMat,]{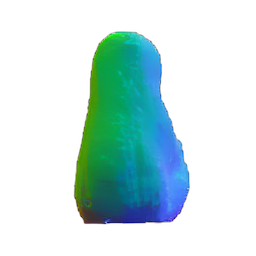}}} &
{\color{black}\fbox{\includegraphics[width=0.8\plotWonderwidthSuppMat]{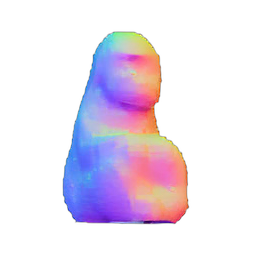}}} &
{\color{black}\fbox{\includegraphics[width=0.8\plotWonderwidthSuppMat,]{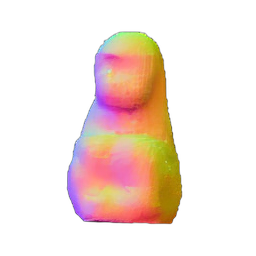}}} &
{\color{black}\fbox{\includegraphics[width=0.8\plotWonderwidthSuppMat]{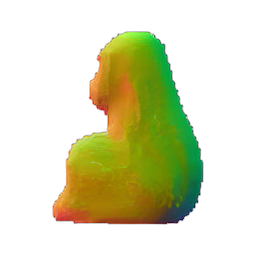}}} & &

{\color{black}\fbox{\includegraphics[width=0.8\plotWonderwidthSuppMat,]{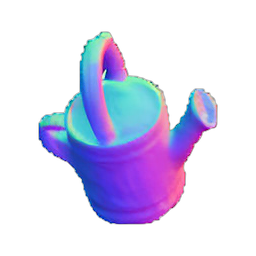}}}&
{\color{black}\fbox{\includegraphics[width=0.8\plotWonderwidthSuppMat]{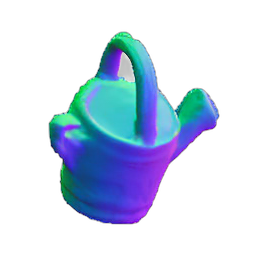}}} &
{\color{black}\fbox{\includegraphics[width=0.8\plotWonderwidthSuppMat,]{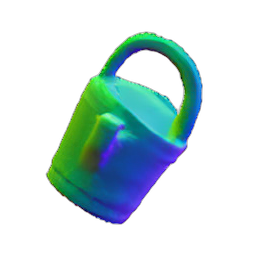}}} &
{\color{black}\fbox{\includegraphics[width=0.8\plotWonderwidthSuppMat]{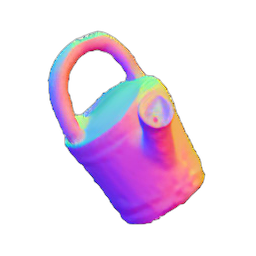}}} &
{\color{black}\fbox{\includegraphics[width=0.8\plotWonderwidthSuppMat,]{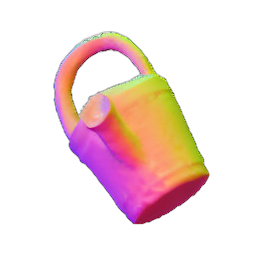}}} &
{\color{black}\fbox{\includegraphics[width=0.8\plotWonderwidthSuppMat]{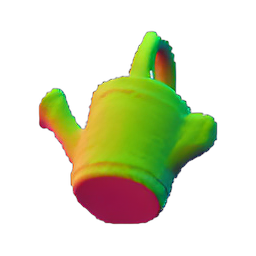}}}\\

\noalign{\vspace{0.5ex}}
 {\color{green}\fbox{\includegraphics[width=0.8\plotWonderwidthSuppMat,]{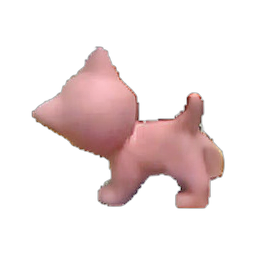}}}&
{\color{black}\fbox{\includegraphics[width=0.8\plotWonderwidthSuppMat]{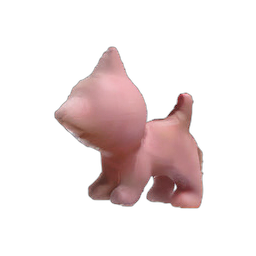}}} &
{\color{black}\fbox{\includegraphics[width=0.8\plotWonderwidthSuppMat,]{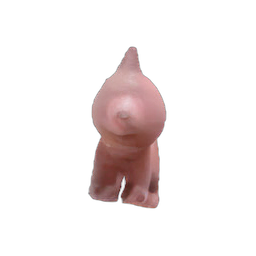}}} &
{\color{black}\fbox{\includegraphics[width=0.8\plotWonderwidthSuppMat]{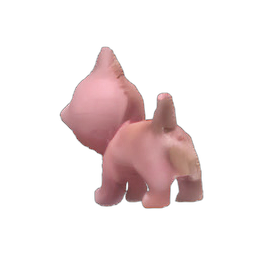}}} &
{\color{black}\fbox{\includegraphics[width=0.8\plotWonderwidthSuppMat,]{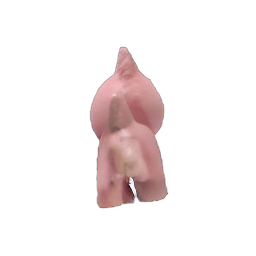}}} &
{\color{black}\fbox{\includegraphics[width=0.8\plotWonderwidthSuppMat]{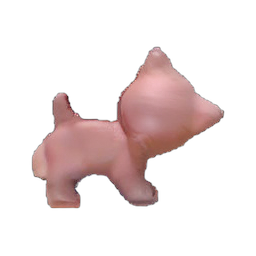}}} & &

{\color{green}\fbox{\includegraphics[width=0.8\plotWonderwidthSuppMat,]{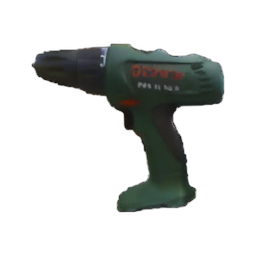}}}&
{\color{black}\fbox{\includegraphics[width=0.8\plotWonderwidthSuppMat]{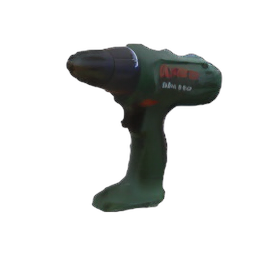}}} &
{\color{black}\fbox{\includegraphics[width=0.8\plotWonderwidthSuppMat,]{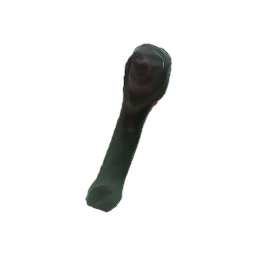}}} &
{\color{black}\fbox{\includegraphics[width=0.8\plotWonderwidthSuppMat]{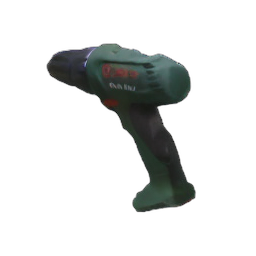}}} &
{\color{black}\fbox{\includegraphics[width=0.8\plotWonderwidthSuppMat,]{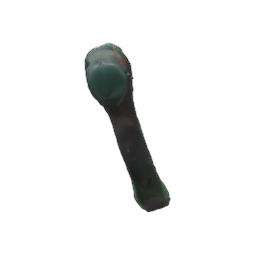}}} &
{\color{black}\fbox{\includegraphics[width=0.8\plotWonderwidthSuppMat]{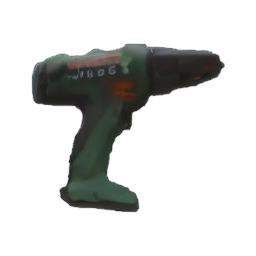}}} \\

\noalign{\vspace{0.5ex}}
{\color{black}\fbox{\includegraphics[width=0.8\plotWonderwidthSuppMat,]{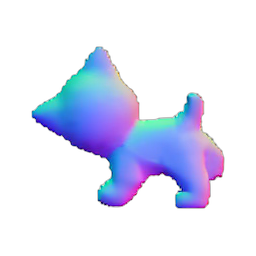}}}&
{\color{black}\fbox{\includegraphics[width=0.8\plotWonderwidthSuppMat]{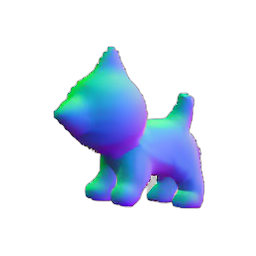}}} &
{\color{black}\fbox{\includegraphics[width=0.8\plotWonderwidthSuppMat,]{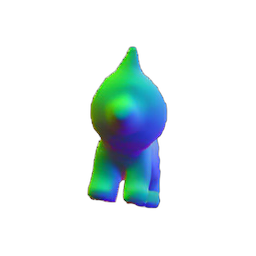}}} &
{\color{black}\fbox{\includegraphics[width=0.8\plotWonderwidthSuppMat]{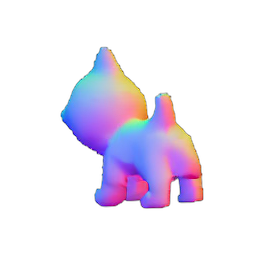}}} &
{\color{black}\fbox{\includegraphics[width=0.8\plotWonderwidthSuppMat,]{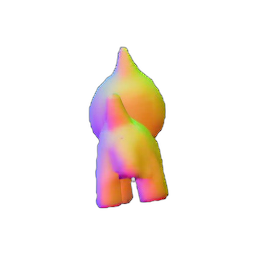}}} &
{\color{black}\fbox{\includegraphics[width=0.8\plotWonderwidthSuppMat]{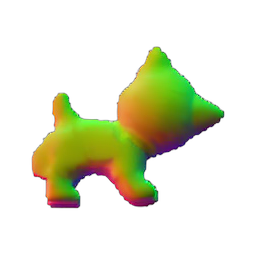}}} & &

{\color{black}\fbox{\includegraphics[width=0.8\plotWonderwidthSuppMat,]{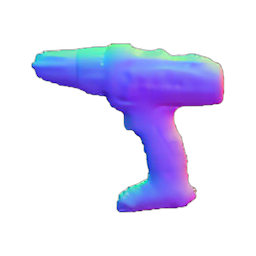}}}&
{\color{black}\fbox{\includegraphics[width=0.8\plotWonderwidthSuppMat]{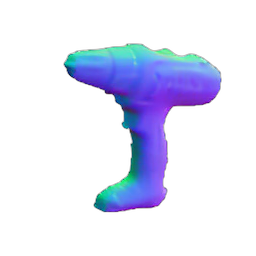}}} &
{\color{black}\fbox{\includegraphics[width=0.8\plotWonderwidthSuppMat,]{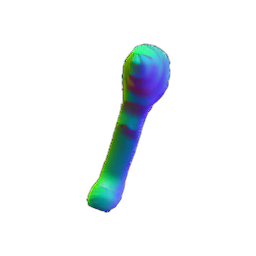}}} &
{\color{black}\fbox{\includegraphics[width=0.8\plotWonderwidthSuppMat]{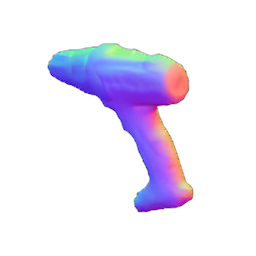}}} &
{\color{black}\fbox{\includegraphics[width=0.8\plotWonderwidthSuppMat,]{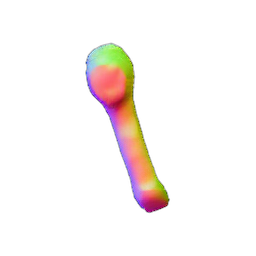}}} &
{\color{black}\fbox{\includegraphics[width=0.8\plotWonderwidthSuppMat]{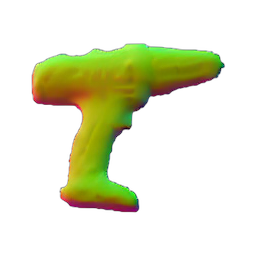}}}\\

\noalign{\vspace{0.5ex}}
{\color{green}\fbox{\includegraphics[width=0.8\plotWonderwidthSuppMat,]{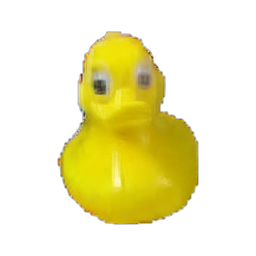}}}&
{\color{black}\fbox{\includegraphics[width=0.8\plotWonderwidthSuppMat]{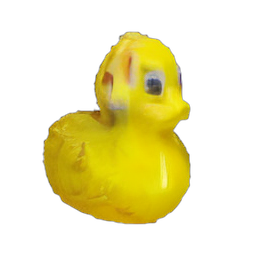}}} &
{\color{black}\fbox{\includegraphics[width=0.8\plotWonderwidthSuppMat,]{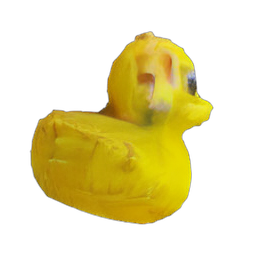}}} &
{\color{black}\fbox{\includegraphics[width=0.8\plotWonderwidthSuppMat]{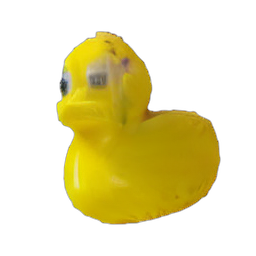}}} &
{\color{black}\fbox{\includegraphics[width=0.8\plotWonderwidthSuppMat,]{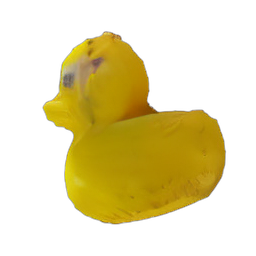}}} &
{\color{black}\fbox{\includegraphics[width=0.8\plotWonderwidthSuppMat]{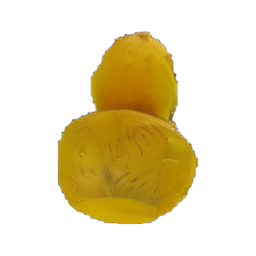}}} & &

{\color{green}\fbox{\includegraphics[width=0.8\plotWonderwidthSuppMat,]{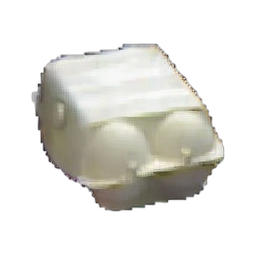}}}&
{\color{black}\fbox{\includegraphics[width=0.8\plotWonderwidthSuppMat]{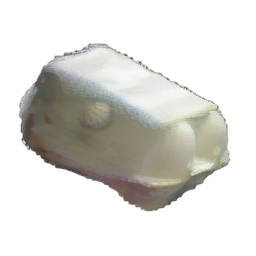}}} &
{\color{black}\fbox{\includegraphics[width=0.8\plotWonderwidthSuppMat,]{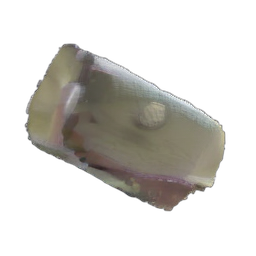}}} &
{\color{black}\fbox{\includegraphics[width=0.8\plotWonderwidthSuppMat]{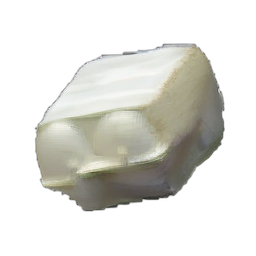}}} &
{\color{black}\fbox{\includegraphics[width=0.8\plotWonderwidthSuppMat,]{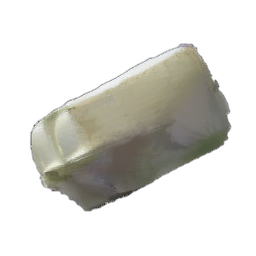}}} &
{\color{black}\fbox{\includegraphics[width=0.8\plotWonderwidthSuppMat]{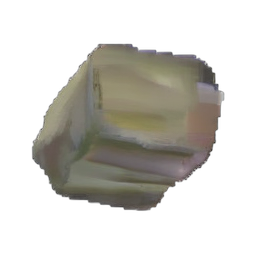}}} \\

\noalign{\vspace{0.5ex}}
{\color{black}\fbox{\includegraphics[width=0.8\plotWonderwidthSuppMat,]{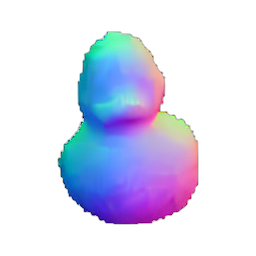}}}&
{\color{black}\fbox{\includegraphics[width=0.8\plotWonderwidthSuppMat]{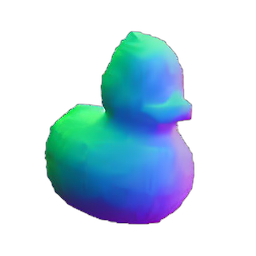}}} &
{\color{black}\fbox{\includegraphics[width=0.8\plotWonderwidthSuppMat,]{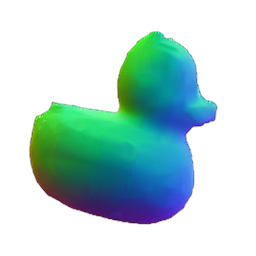}}} &
{\color{black}\fbox{\includegraphics[width=0.8\plotWonderwidthSuppMat]{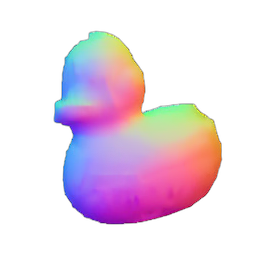}}} &
{\color{black}\fbox{\includegraphics[width=0.8\plotWonderwidthSuppMat,]{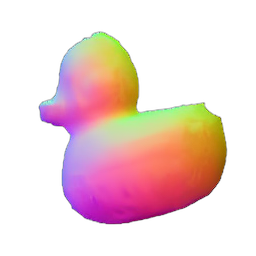}}} &
{\color{black}\fbox{\includegraphics[width=0.8\plotWonderwidthSuppMat]{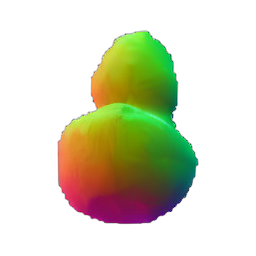}}} & &

{\color{black}\fbox{\includegraphics[width=0.8\plotWonderwidthSuppMat,]{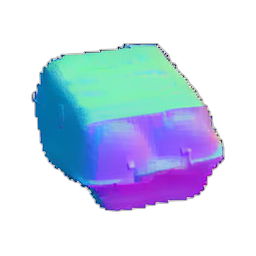}}}&
{\color{black}\fbox{\includegraphics[width=0.8\plotWonderwidthSuppMat]{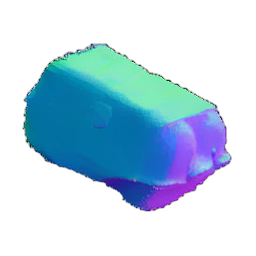}}} &
{\color{black}\fbox{\includegraphics[width=0.8\plotWonderwidthSuppMat,]{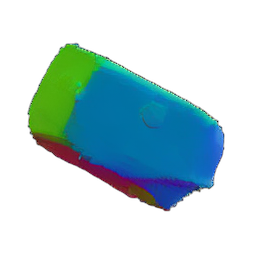}}} &
{\color{black}\fbox{\includegraphics[width=0.8\plotWonderwidthSuppMat]{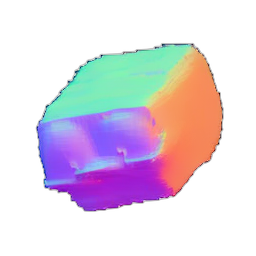}}} &
{\color{black}\fbox{\includegraphics[width=0.8\plotWonderwidthSuppMat,]{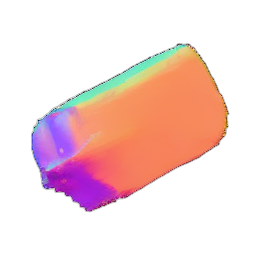}}} &
{\color{black}\fbox{\includegraphics[width=0.8\plotWonderwidthSuppMat]{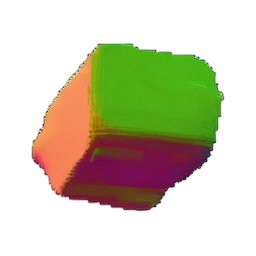}}}\\

\noalign{\vspace{0.5ex}}
{\color{green}\fbox{\includegraphics[width=0.8\plotWonderwidthSuppMat,]{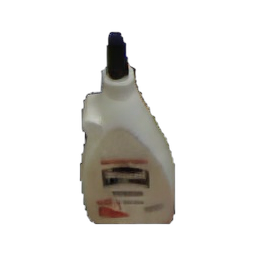}}}&
{\color{black}\fbox{\includegraphics[width=0.8\plotWonderwidthSuppMat]{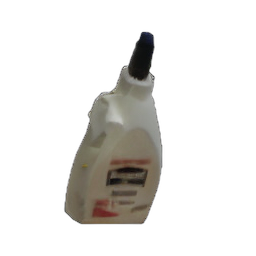}}} &
{\color{black}\fbox{\includegraphics[width=0.8\plotWonderwidthSuppMat,]{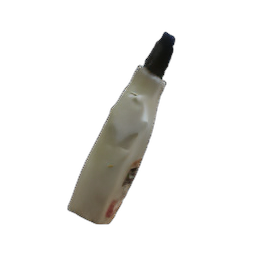}}} &
{\color{black}\fbox{\includegraphics[width=0.8\plotWonderwidthSuppMat]{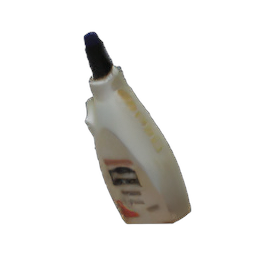}}} &
{\color{black}\fbox{\includegraphics[width=0.8\plotWonderwidthSuppMat,]{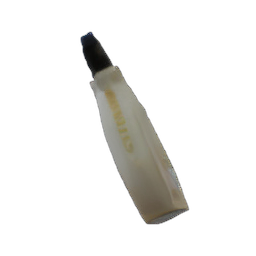}}} &
{\color{black}\fbox{\includegraphics[width=0.8\plotWonderwidthSuppMat]{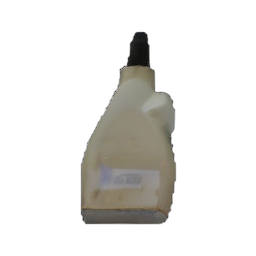}}} & &

{\color{green}\fbox{\includegraphics[width=0.8\plotWonderwidthSuppMat,]{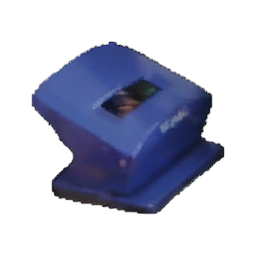}}}&
{\color{black}\fbox{\includegraphics[width=0.8\plotWonderwidthSuppMat]{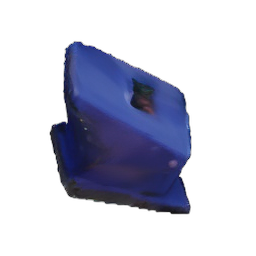}}} &
{\color{black}\fbox{\includegraphics[width=0.8\plotWonderwidthSuppMat,]{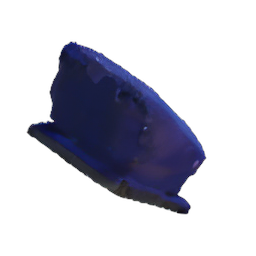}}} &
{\color{black}\fbox{\includegraphics[width=0.8\plotWonderwidthSuppMat]{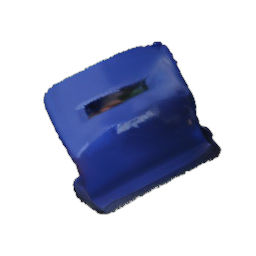}}} &
{\color{black}\fbox{\includegraphics[width=0.8\plotWonderwidthSuppMat,]{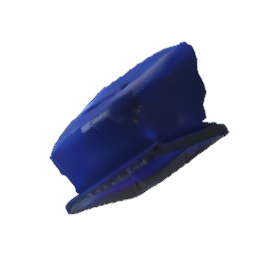}}} &
{\color{black}\fbox{\includegraphics[width=0.8\plotWonderwidthSuppMat]{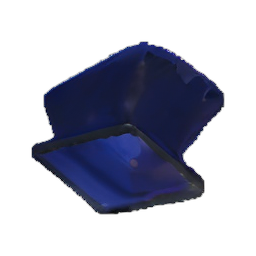}}} \\

\noalign{\vspace{0.5ex}}
{\color{black}\fbox{\includegraphics[width=0.8\plotWonderwidthSuppMat,]{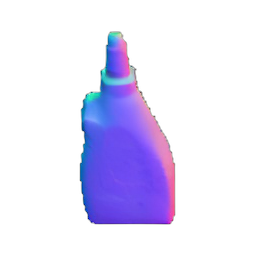}}}&
{\color{black}\fbox{\includegraphics[width=0.8\plotWonderwidthSuppMat]{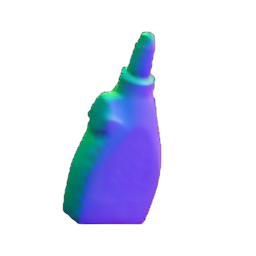}}} &
{\color{black}\fbox{\includegraphics[width=0.8\plotWonderwidthSuppMat,]{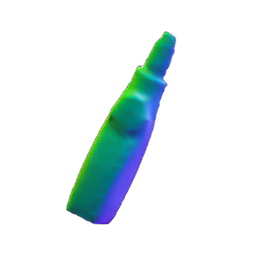}}} &
{\color{black}\fbox{\includegraphics[width=0.8\plotWonderwidthSuppMat]{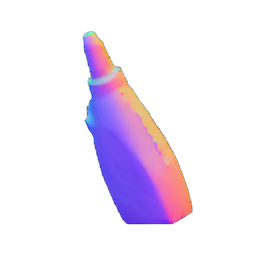}}} &
{\color{black}\fbox{\includegraphics[width=0.8\plotWonderwidthSuppMat,]{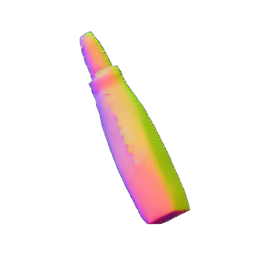}}} &
{\color{black}\fbox{\includegraphics[width=0.8\plotWonderwidthSuppMat]{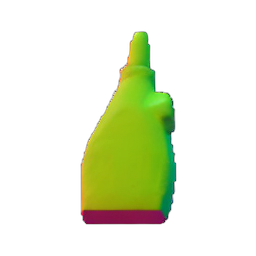}}} & &

{\color{black}\fbox{\includegraphics[width=0.8\plotWonderwidthSuppMat]{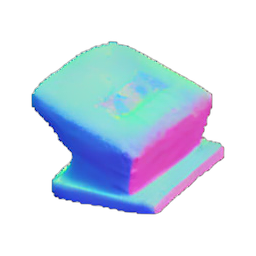}}}&
{\color{black}\fbox{\includegraphics[width=0.8\plotWonderwidthSuppMat]{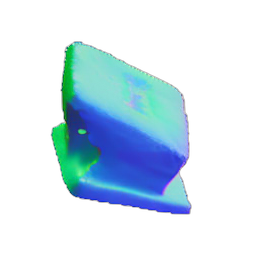}}} &
{\color{black}\fbox{\includegraphics[width=0.8\plotWonderwidthSuppMat,]{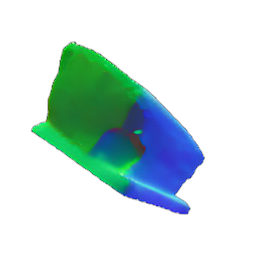}}} &
{\color{black}\fbox{\includegraphics[width=0.8\plotWonderwidthSuppMat]{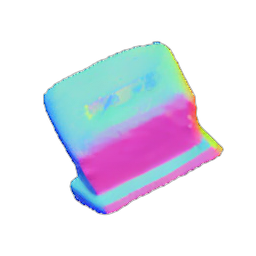}}} &
{\color{black}\fbox{\includegraphics[width=0.8\plotWonderwidthSuppMat,]{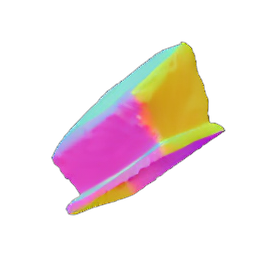}}} &
{\color{black}\fbox{\includegraphics[width=0.8\plotWonderwidthSuppMat]{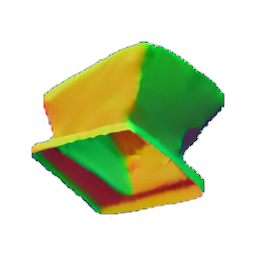}}}\\

\scriptsize{Front} & \scriptsize{Front left} & \scriptsize{Left} & \scriptsize{Front-right} & \scriptsize{Right} & \scriptsize{Back} & & \scriptsize{Front} & \scriptsize{Front left} & \scriptsize{Left} & \scriptsize{Front-right} & \scriptsize{Right} & \scriptsize{Back}
\end{tabular}
}

    \captionof{figure}{ 
     \textbf{3D reconstruction by Wonder3D~\cite{long2023wonder3d}}.
     For each sample (2$\times$6 images), we show the input image outlined in {\color{darkgreen}{green}} (top left), the predicted rgb (second to last column of the first row), and the second row shows the corresponding predicted normals.}
    \label{fig:wonder3d}
    \vspace*{-5pt}
\end{center}
\end{figure*}

As discussed in Section 4 of the main paper, due to the sensitivity of Wonder3D to the quality of input images, we selected reference images from the test set of LM~\cite{hinterstoisser-accv12-modelbasedtrainingdetection} based on three criteria: (i) not present in the test set of LM-O~\cite{brachmann-eccv14-learning6dobjectposeestimation}, (ii) the target object is fully visible and (iii) well segmented by Segment Anything~\cite{kirillov2023segment}. Despite these careful selections, we observe that Wonder3D can still fail due to the low resolution or the lack of ``perspective'' information in the input image. Figure~\ref{fig:failureCases} illustrates these common failure cases of Wonder3D where objects appear ``flat'' when viewed from novel angles.

We therefore select reference images for each object, sourced from ``scene\_id/im\_id'' images as follows (sorted by ``object\_id''): ``000001/000693'', ``000005/000775'', ``000006/000949'', ``000008/000994'', ``000009/001228'', ``000010/000289'', ``000011/001069'', and ``000012/000647''. Figure~\ref{fig:wonder3d} presents additional visualizations of 3D reconstruction from a single image by Wonder3D~\cite{long2023wonder3d}, using these images. It is important to note that the final 3D models are reconstructed from these six views using the instant-NGP based SDF reconstruction method~\cite{instant-nsr-pl}. The reference is defined as the front view, while the five predicted views are the front-left, left, front-right, right, and back.

We show in Table~\ref{tab:cnos} CNOS's results for both detection and segmentation tasks when using 3D models predicted from a single image by Wonder3D~\cite{long2023wonder3d}.

\begin{table}[t]
\centering
\setlength{\tabcolsep}{2.5pt}
\scalebox{0.9}{
\begin{tabular}{cccc}

\toprule  
 Input  & Detection  & Segmentation \\ 

\midrule  %
 GT 3D model & 43.3  & 39.7\\ %
 Predicted 3D model & 38.3  & 36.3\\ %
\bottomrule

\end{tabular}
}
\vspace{-5pt}

\caption{{\bf CNOS~\cite{nguyen2023cnos}'s performance on LM-O dataset~\cite{brachmann-eccv14-learning6dobjectposeestimation}.} For this evaluation, we use the standard protocol designed for detection and segmentation tasks, used in Tasks 5 and 6 of the BOP Challenge of BOP challenge~\cite{sundermeyer2023bop}.}
\label{tab:cnos}
\end{table}

\subsection{Using ground-truth 3D models}
We show in Figure~\ref{fig:lmo}, and~\ref{fig:ycbv} qualitative results on challenging conditions of LM-O~\cite{brachmann-eccv14-learning6dobjectposeestimation} and YCB-V~\cite{Xiang2018-dv} datasets.  

\section{Future work}
As discussed in Section 4.3 of the main paper, GigaPose fails on challenging conditions where heavy occlusions, low-resolution segmentation, and low-fidelity CAD models are present, as observed in the LM-O dataset. To address this issue, incorporating additional modalities, such as depth images, can be beneficial. Depth images, which capture information about object geometries, can significantly enhance model performance under these conditions. Moreover, although our method shows promising results in a single-reference setting using Wonder3D~\cite{long2023wonder3d}, it requires manual selection to achieve high-quality 3D reconstruction. Therefore, the development of more advanced 3D reconstruction techniques capable of generating high-quality outputs from a single image would be particularly valuable in this context.

\newpage
\newlength{\plotwidthSuppMat}
\setlength\plotwidthSuppMat{2.2cm}
\setlength\lineskip{10pt}
\setlength\tabcolsep{2pt} 

\begin{figure*}[!h]
\begin{center}
{\small
\begin{tabular}{
>{\centering\arraybackslash}m{2cm}
>{\centering\arraybackslash}m{\plotwidthSuppMat}
>{\centering\arraybackslash}m{\plotwidthSuppMat}
>{\centering\arraybackslash}m{2\plotwidthSuppMat}
>{\centering\arraybackslash}m{\plotwidthSuppMat}
>{\centering\arraybackslash}m{\plotwidthSuppMat}
}
& zoom-out& & & & zoom-out\\

Ape & \frame{\includegraphics[height=\plotwidthSuppMat,]{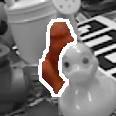}}&
\frame{\includegraphics[height=\plotwidthSuppMat,]{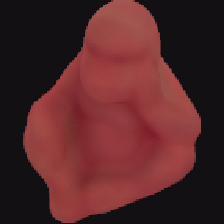}}&
\frame{\includegraphics[height=\plotwidthSuppMat, ]{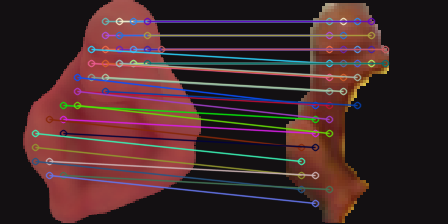}} &
\frame{\includegraphics[height=\plotwidthSuppMat,]{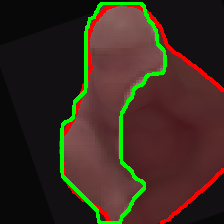}} &
\frame{\includegraphics[height=\plotwidthSuppMat, ]{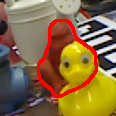}}\\

Watering\_can & \frame{\includegraphics[height=\plotwidthSuppMat,]{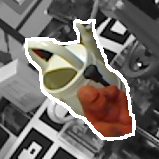}}&
\frame{\includegraphics[height=\plotwidthSuppMat,]{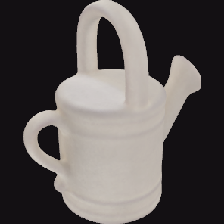}}&
\frame{\includegraphics[height=\plotwidthSuppMat, ]{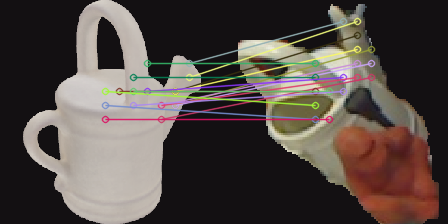}} &
\frame{\includegraphics[height=\plotwidthSuppMat,]{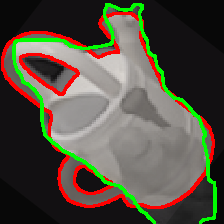}} &
\frame{\includegraphics[height=\plotwidthSuppMat, ]{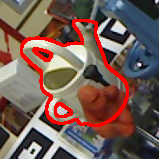}}\\

Cat & 
 \frame{\includegraphics[height=\plotwidthSuppMat,]{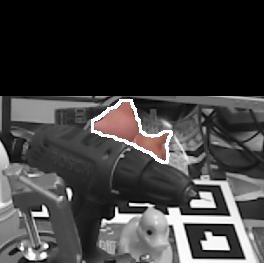}}&
\frame{\includegraphics[height=\plotwidthSuppMat,]{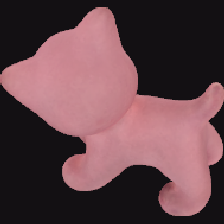}}&
\frame{\includegraphics[height=\plotwidthSuppMat, ]{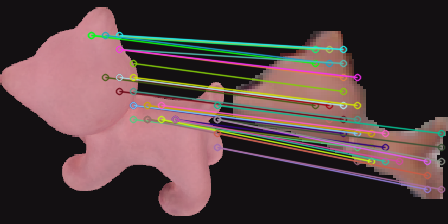}} &
\frame{\includegraphics[height=\plotwidthSuppMat,]{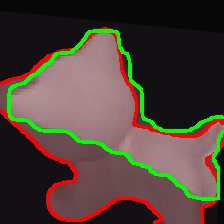}} &
\frame{\includegraphics[height=\plotwidthSuppMat, ]{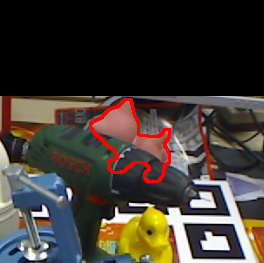}}\\

Drill & \frame{\includegraphics[height=\plotwidthSuppMat,]{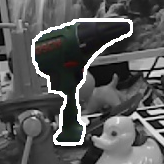}}&
\frame{\includegraphics[height=\plotwidthSuppMat,]{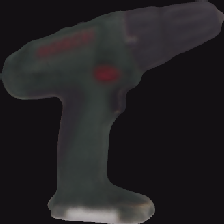}}&
\frame{\includegraphics[height=\plotwidthSuppMat, ]{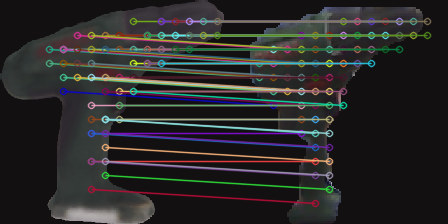}} &
\frame{\includegraphics[height=\plotwidthSuppMat,]{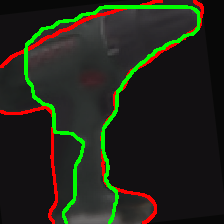}} &
\frame{\includegraphics[height=\plotwidthSuppMat, ]{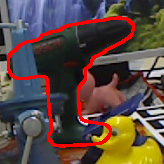}}\\

Duck & \frame{\includegraphics[height=\plotwidthSuppMat,]{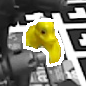}}&
\frame{\includegraphics[height=\plotwidthSuppMat,]{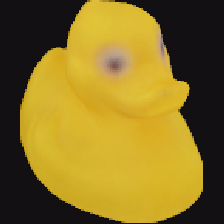}}&
\frame{\includegraphics[height=\plotwidthSuppMat, ]{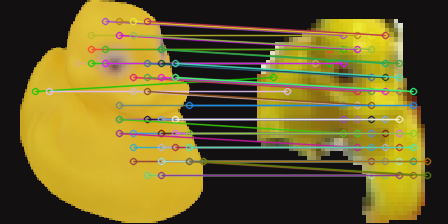}} &
\frame{\includegraphics[height=\plotwidthSuppMat,]{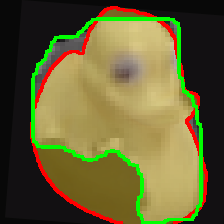}} &
\frame{\includegraphics[height=\plotwidthSuppMat, ]{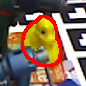}}\\

Eggbox & \frame{\includegraphics[height=\plotwidthSuppMat,]{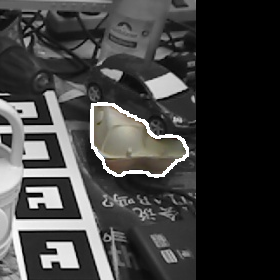}}&
\frame{\includegraphics[height=\plotwidthSuppMat,]{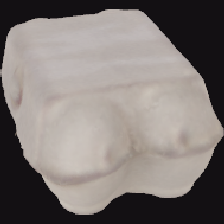}}&
\frame{\includegraphics[height=\plotwidthSuppMat, ]{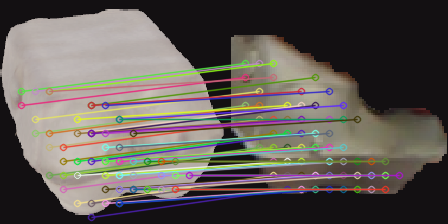}} &
\frame{\includegraphics[height=\plotwidthSuppMat,]{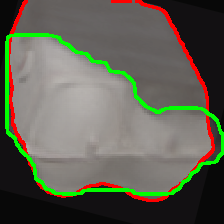}} &
\frame{\includegraphics[height=\plotwidthSuppMat, ]{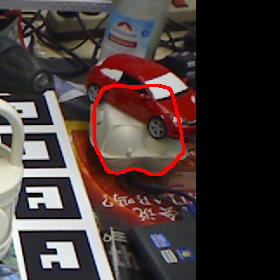}}\\

Glue & \frame{\includegraphics[height=\plotwidthSuppMat,]{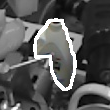}}&
\frame{\includegraphics[height=\plotwidthSuppMat,]{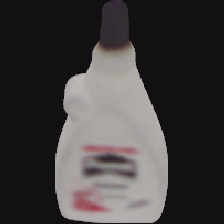}}&
\frame{\includegraphics[height=\plotwidthSuppMat, ]{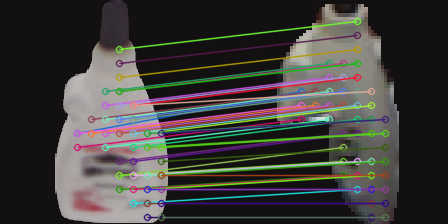}} &
\frame{\includegraphics[height=\plotwidthSuppMat,]{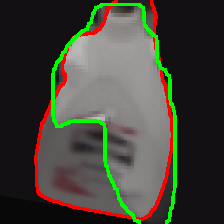}} &
\frame{\includegraphics[height=\plotwidthSuppMat, ]{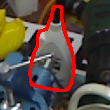}}\\

Hole\_puncher & \frame{\includegraphics[height=\plotwidthSuppMat,]{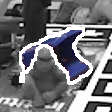}}&
\frame{\includegraphics[height=\plotwidthSuppMat,]{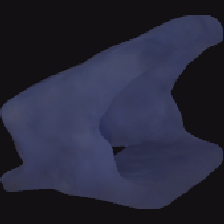}}&
\frame{\includegraphics[height=\plotwidthSuppMat, ]{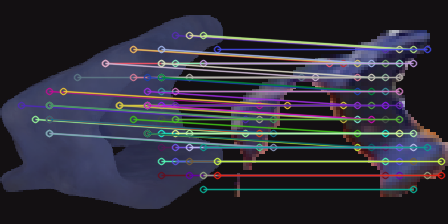}} &
\frame{\includegraphics[height=\plotwidthSuppMat,]{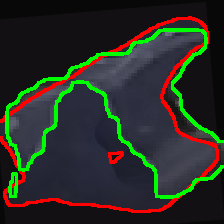}} &
\frame{\includegraphics[height=\plotwidthSuppMat, ]{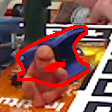}}\\ 

&  \multirow{2}{*}{{\makecell[b]{Input \\ segmentation~\cite{nguyen2023cnos}}}}  & \multirow{2}{*}{{\makecell[b]{Nearest \\ template ($\Routplane$)}}}   &  \multirow{2}{*}{{\makecell[b]{2D-to-2D correspondences}}} & \multirow{2}{*}{{\makecell[b]{Alignment $\affineTransform$}}}  & \multirow{2}{*}{{\makecell[b]{Final prediction}}} \\

\end{tabular}
}
\vspace{20pt}
    \captionof{figure}{ 
     \textbf{Qualitative results on LM-O~\cite{brachmann-eccv14-learning6dobjectposeestimation}}. The first column shows CNOS~\cite{nguyen2023cnos}'s segmentation. The second and third columns illustrate the outputs of the nearest neighbor search step, which includes the nearest template ($\Routplane$) and the 2D-to-2D correspondences. The fourth column demonstrates the alignment achieved by applying the predicted affine transform $\affineTransform$ to the template, then overlaying it on the query input: The {\color{darkgreen}{green}} contour indicates the noisy segmentation by CNOS~\cite{nguyen2023cnos}, while the {\color{red}{red}} contour highlights the boundary of the aligned template. The last column show the final prediction after refinement~\cite{megapose}.}
    \label{fig:lmo}
    \vspace*{-5pt}
\end{center}
\end{figure*}
\newpage
\setlength\plotwidth{2.2cm}
\setlength\lineskip{10pt}
\setlength\tabcolsep{2pt} 

\begin{figure*}[!h]
\begin{center}
{\small
\begin{tabular}{
>{\centering\arraybackslash}m{2cm}
>{\centering\arraybackslash}m{\plotwidth}
>{\centering\arraybackslash}m{\plotwidth}
>{\centering\arraybackslash}m{2\plotwidth}
>{\centering\arraybackslash}m{\plotwidth}
>{\centering\arraybackslash}m{\plotwidth}
}
& zoom-out& & & & zoom-out\\

Object ``02'' & \frame{\includegraphics[height=\plotwidth,]{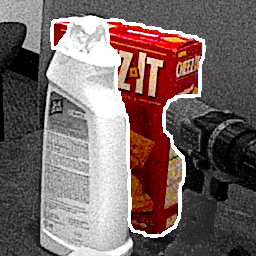}}&
\frame{\includegraphics[height=\plotwidth,]{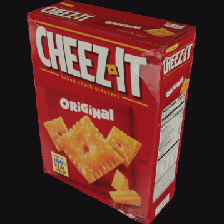}}&
\frame{\includegraphics[height=\plotwidth, ]{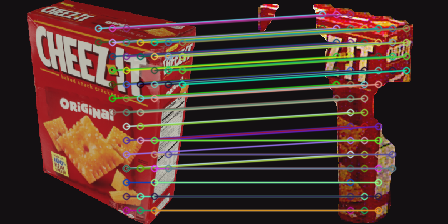}} &
\frame{\includegraphics[height=\plotwidth,]{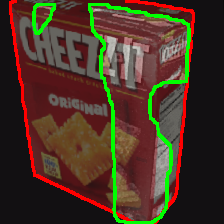}} &
\frame{\includegraphics[height=\plotwidth, ]{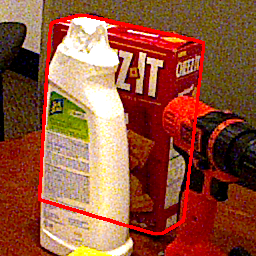}}\\

Object ``03'' & \frame{\includegraphics[height=\plotwidth,]{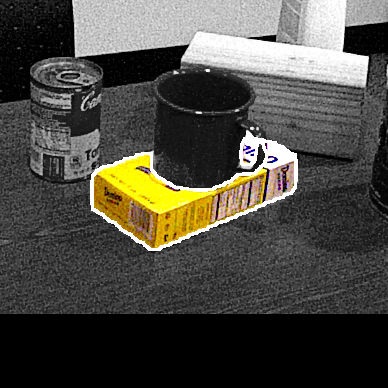}}&
\frame{\includegraphics[height=\plotwidth,]{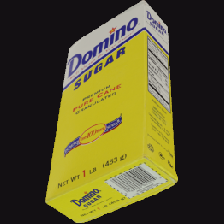}}&
\frame{\includegraphics[height=\plotwidth, ]{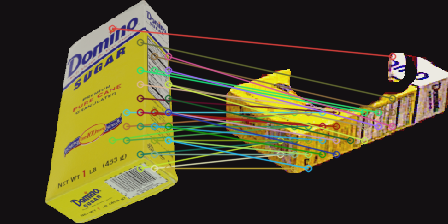}} &
\frame{\includegraphics[height=\plotwidth,]{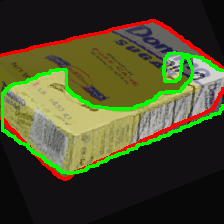}} &
\frame{\includegraphics[height=\plotwidth, ]{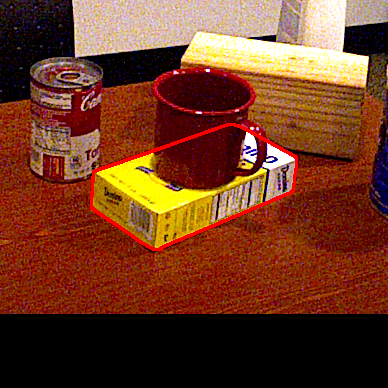}}\\

Object ``04'' & \frame{\includegraphics[height=\plotwidth,]{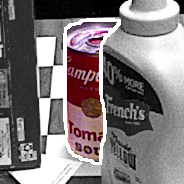}}&
\frame{\includegraphics[height=\plotwidth,]{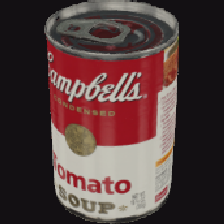}}&
\frame{\includegraphics[height=\plotwidth, ]{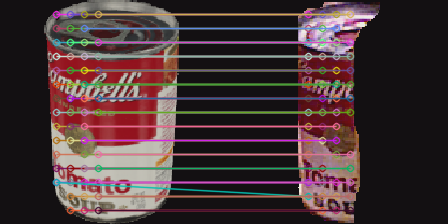}} &
\frame{\includegraphics[height=\plotwidth,]{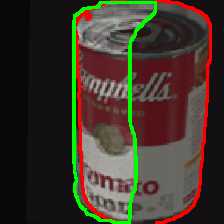}} &
\frame{\includegraphics[height=\plotwidth, ]{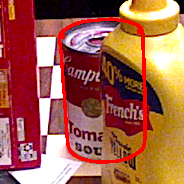}}\\

Object ``10'' & \frame{\includegraphics[height=\plotwidth,]{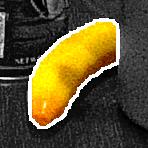}}&
\frame{\includegraphics[height=\plotwidth,]{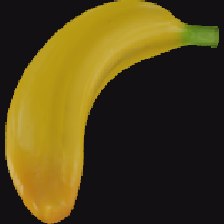}}&
\frame{\includegraphics[height=\plotwidth, ]{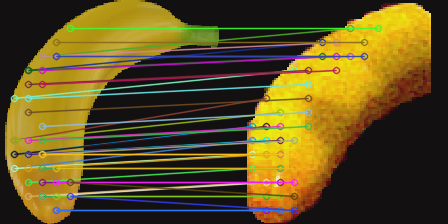}} &
\frame{\includegraphics[height=\plotwidth,]{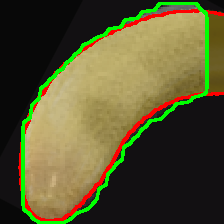}} &
\frame{\includegraphics[height=\plotwidth, ]{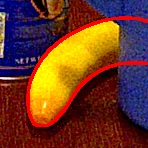}}\\

Object ``13'' & \frame{\includegraphics[height=\plotwidth,]{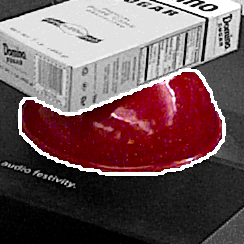}}&
\frame{\includegraphics[height=\plotwidth,]{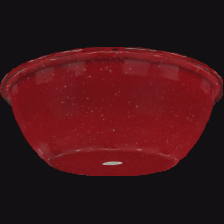}}&
\frame{\includegraphics[height=\plotwidth, ]{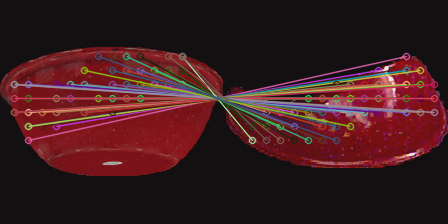}} &
\frame{\includegraphics[height=\plotwidth,]{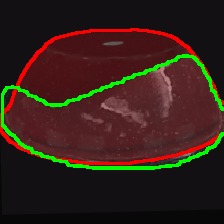}} &
\frame{\includegraphics[height=\plotwidth, ]{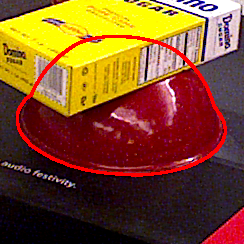}}\\

Object ``15'' & \frame{\includegraphics[height=\plotwidth,]{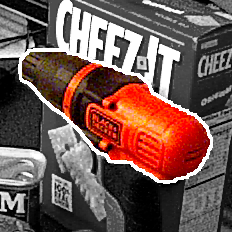}}&
\frame{\includegraphics[height=\plotwidth,]{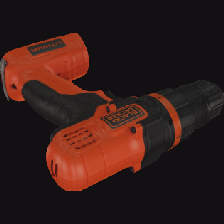}}&
\frame{\includegraphics[height=\plotwidth, ]{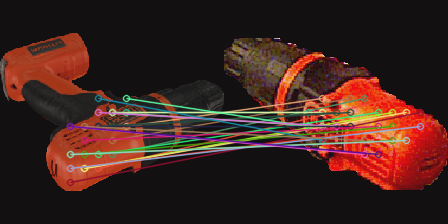}} &
\frame{\includegraphics[height=\plotwidth,]{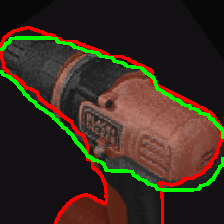}} &
\frame{\includegraphics[height=\plotwidth, ]{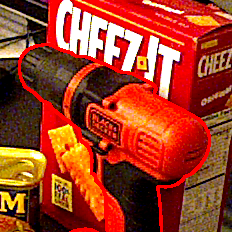}}\\

Object ``16'' & \frame{\includegraphics[height=\plotwidth,]{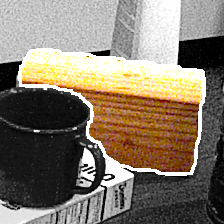}}&
\frame{\includegraphics[height=\plotwidth,]{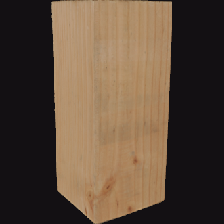}}&
\frame{\includegraphics[height=\plotwidth, ]{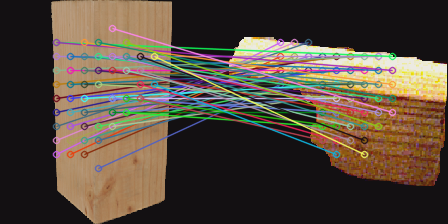}} &
\frame{\includegraphics[height=\plotwidth,]{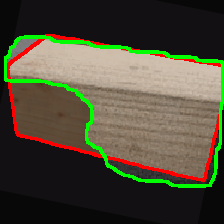}} &
\frame{\includegraphics[height=\plotwidth, ]{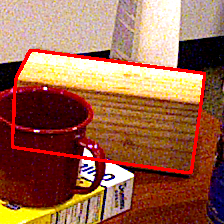}}\\

Object ``21'' & \frame{\includegraphics[height=\plotwidth,]{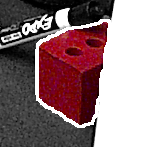}}&
\frame{\includegraphics[height=\plotwidth,]{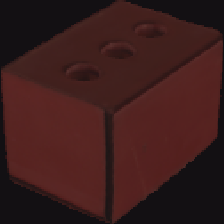}}&
\frame{\includegraphics[height=\plotwidth, ]{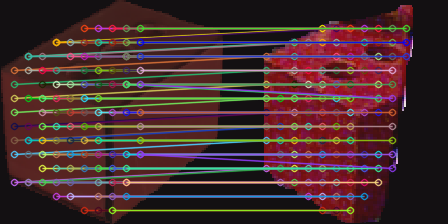}} &
\frame{\includegraphics[height=\plotwidth,]{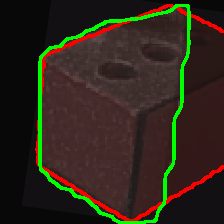}} &
\frame{\includegraphics[height=\plotwidth, ]{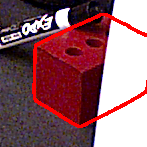}}\\ 

&  \multirow{2}{*}{{\makecell[b]{Input \\ segmentation~\cite{nguyen2023cnos}}}}  & \multirow{2}{*}{{\makecell[b]{Nearest \\ template ($\Routplane$)}}}   &  \multirow{2}{*}{{\makecell[b]{2D-to-2D correspondences}}} & \multirow{2}{*}{{\makecell[b]{Alignment $\affineTransform$}}}  & \multirow{2}{*}{{\makecell[b]{Final prediction}}} \\

\end{tabular}
}
\vspace{20pt}
    \captionof{figure}{ 
     \textbf{Qualitative results on YCB-V~\cite{Xiang2018-dv}}. The first column shows CNOS~\cite{nguyen2023cnos}'s segmentation. The second and third columns illustrate the outputs of the nearest neighbor search step, which includes the nearest template ($\Routplane$) and the 2D-to-2D correspondences. The fourth column demonstrates the alignment achieved by applying the predicted affine transform $\affineTransform$ to the template, then overlaying it on the query input: The {\color{darkgreen}{green}} contour indicates the noisy segmentation by CNOS~\cite{nguyen2023cnos}, while the {\color{red}{red}} contour highlights the boundary of the aligned template. The last column show the final prediction after refinement~\cite{megapose}.}
    \label{fig:ycbv}
\end{center}
\end{figure*}




 \clearpage
{\small \noindent\textbf{Acknowledgments.} The authors extend their gratitude to Jonathan Tremblay for sharing the visualizations of DiffDOPE and providing valuable feedback. We also thank Médéric Fourmy and Sungphill Moon for sharing the results of MegaPose and of GenFlow in the BOP Challenge 2023, and Yann Labbé for allowing the authors to use the name GigaPose. The authors thank Micha\"el Ramamonjisoa and Constantin Aronssohn for helpful discussions. This project was funded in part by the European Union (ERC Advanced Grant explorer  Funding ID \#101097259). This work was performed using HPC resources from GENCI–IDRIS 2022-AD011012294R2.
{\small
\bibliographystyle{ieee_fullname}
\bibliography{cleaned_refs}
}

\end{document}